\documentclass[11pt]{article}
\usepackage{kpfonts}
\usepackage{fullpage}
\usepackage{hyperref}       
\usepackage{url}            
\usepackage{booktabs}       
\usepackage{amsfonts}       
\usepackage{nicefrac}       
\usepackage{amsmath}
\usepackage{amssymb}
\usepackage{amsthm}
\usepackage{natbib}
\usepackage{multirow}
\usepackage{graphics}
\usepackage{array}

\DeclareMathOperator*{\argmin}{arg\,min}

\newcommand{\xhdr}[1]{\vspace{2mm} \noindent{\bf #1}}

\newcommand{\clip}{\text{clip}}

\newcommand{\lla}{\left\langle}

\newcommand{\rra}{\right\rangle}

\usepackage{tcolorbox}

\newtheorem{theorem}{Theorem}
\newtheorem{corollary}{Corollary}
\newtheorem{lemma}{Lemma}

\makeatletter
\newtheorem*{rep@theorem}{\rep@title}
\newcommand{\newreptheorem}[2]{%
\newenvironment{rep#1}[1]{%
 \def\rep@title{#2 \ref{##1}}%
 \begin{rep@theorem}}%
 {\end{rep@theorem}}}
\makeatother
\newreptheorem{theorem}{Theorem}

\title{Understanding Gradient Clipping in Private SGD: \\A Geometric Perspective}
\author{%
  Xiangyi Chen\thanks{ University of Minnesota.
  {\texttt {email:chen5719@umn.edu}}} 
  \and 
  Zhiwei Steven Wu\thanks{Carnegie Mellon University.
  {\texttt {email:zstevenwu@cmu.edu}}}
  \and 
  Mingyi Hong\thanks{ University of Minnesota. {\texttt {email:mhong@umn.edu}}
}
}

\date{}

\begin{document}

\maketitle
\begin{abstract}
Deep learning models are increasingly popular in many machine learning applications where the training data may contain sensitive information. To provide formal and rigorous privacy guarantee, many learning systems now incorporate differential privacy by training their models with \emph{(differentially) private SGD}. A key step in each private SGD update is \emph{gradient clipping} that shrinks the gradient of an individual example whenever its $\ell_2$ norm exceeds some threshold. We first demonstrate how gradient clipping can prevent SGD from converging to a stationary point. We then provide a theoretical analysis that fully quantifies the clipping bias on convergence with a disparity measure between the gradient distribution and a geometrically symmetric distribution. Our empirical evaluation further suggests that the gradient distributions along the trajectory of private SGD indeed exhibit symmetric structure that favors convergence. Together, our results provide an explanation why private SGD with gradient clipping remains effective in practice despite its potential clipping bias. {Finally, we develop a new perturbation-based technique that can provably correct the clipping bias even for instances with highly asymmetric gradient distributions.}
\end{abstract}

\section{Introduction}\label{sec:intro}

Many modern applications of machine learning rely on datasets that may contain sensitive personal information, including medical records, browsing history, and geographic locations. To protect the private information of individual citizens, many machine learning systems now train their models subject to the constraint of differential privacy~\citep{DMNS06}, which informally requires that no individual training example has a significant influence on the trained model. To achieve this formal privacy guarantee, one of the most popular training methods, especially for deep learning, is \emph{differentially private stochastic gradient descent} (DP-SGD)~\citep{basm14, abadi2016deep}. At a high level, DP-SGD is a simple modification of SGD that makes each step differentially private with the \emph{Gaussian mechanism}: at each iteration $t$, it first computes a gradient estimate $g_t$ based on a random subsample, and then updates the model using a noisy gradient $\tilde g_t = g_t + \eta$, where $\eta$ is a noise vector drawn from a multivariate Gaussian distribution.

Despite the simple form of DP-SGD, there is a major disparity between its theoretical analysis and practical implementation. The formal privacy guarantee of Gaussian mechanism requires that the per-coordinate standard deviation of the noise vector $\eta$ scales linearly with the $\ell_2$ sensitivity of the gradient estimate $g_t$---that is, the maximal change on $g_t$ in $\ell_2$ distance if by changining a single example. To bound the $\ell_2$-sensitivity, existing theoretical analyses typically assume that the loss function is $L$-Lipschitz in the model parameters, and the constant $L$ is known to the algorithm designer for setting the noise rate~\citep{basm14, Wang019}. Since this assumption implies that the gradient of each example has $\ell_2$ norm bounded by $L$, any gradient estimate from averaging over the gradients of $m$ examples has $\ell_2$-sensitivity bounded by $L/m$. However, in many practical settings, especially those with deep learning models, such Lipschitz constant or gradient bounds are not a-priori known or even computable (since it involves taking the worst case over both examples and pairs of parameters). In practice, the bounded $\ell_2$-sensitivity is ensured by \emph{gradient clipping}~\citep{abadi2016deep} that shrinks an individual gradient whenever its $\ell_2$ norm exceeds certain threshold $c$. More formally, given any gradient $g$ on a simple example and a clipping threshold $c$, the gradient clipping does the following
\begin{align}\label{eq: clip}
   & \clip(g,c) = g \cdot \min\left (1,\frac{c}{\|g\|} \right).
\end{align}
However, the clipping operation can create a substantial bias in the update direction. To illustrate this clipping bias, consider the following two optimization problems even without the privacy constraint.

\xhdr{Example 1.} Consider optimizing $f(x) = \frac{1}{3} \sum_{i=1} ^3 \frac{1}{2} (x-a_i)^2$ over $x\in \mathbb{R}$, where $a_1 = a_2 =  -3$ and $a_3 = 9$. Since the gradient $\nabla f(x) = x-1$, the optimum is $x^* = 1$. Now suppose we run SGD with gradient clipping with a threshold of $c = 1$. At the optimum, the gradients for all three examples are clipped and the expected clipped gradient is $1/3$, which leads the parameter to move away from $x^*$.

\xhdr{Example 2.} Let $f(x) = \frac{1}{2} \sum_{i=1} ^2 \frac{1}{2} (x-a_i)^2$,  where $a_1 =  -3$ and $a_2 = 3$.  The minimum of $f$ is achieved at $x^* = 0$, where the expected clipped gradient is also 0. However, SGD with clipped gradients and $c = 1$ may never converge to $x^*$ since the expected clipped gradients are all 0 for any $x\in [-2,2]$, which  means all these points are "stationary" for the algorithm.

Both examples above show that clipping bias can prevent convergence in the worst case. Existing analyses on gradient clipping quantify this clipping bias either with 1) the difference between clipped and unclipped gradients \citep{pichapati2019adaclip}, or 2) the fraction of examples with gradient norms exceeding the clip threshold $c$ \citep{zhang2019gradient}. These approaches suggest that a small clip threshold will lead to large clipping bias and worsen the training performance of DP-SGD. However, in practice, DP-SGD often remains effective even with a small clip threshold, which indicates a gap in the current theoretical understanding of gradient clipping.

\subsection{Our results}
 We study the effects of gradient clipping on SGD and DP-SGD and provide:

\textbf{Symmetricity-based analysis.}
We characterize the clipping bias on the convergence to stationary points through the geometric structure of the gradient distribution. To isolate the clipping effects, we first analyze the non-private SGD with gradient clipping (but without Gaussian perturbation), with the following key analysis steps. 
We first show that the inner product $\mathbb{E}[\langle \nabla f(x_t), g_t\rangle]$ goes to zero in SGD, where $\nabla f(x)$ denotes the true gradient and $g_t$ denotes a clipped stochastic gradient. Secondly, we show that when the gradient distribution is symmetric, the inner product $\mathbb{E}[\langle \nabla f(x_t), g_t\rangle]$ upper bounds a constant re-scaling of $\|\nabla f(x_t)\|$, and so SGD  minimizes the gradient norm. 
We then quantify the clipping bias via a coupling between the gradient distribution and a nearby symmetric distribution and express it as a disparity measure (that resembles the Wasserstein distance) between the two distributions. As a result, when the gradient distributions are near-symmetric or when the clipping bias favors convergence, the clipped gradient remains aligned with the true gradient, even if clipping aggressively shrinks almost all the {sample} gradients.

\textbf{Theoretical and empirical evaluation of DP-SGD.} 
 Building on the SGD analysis, we obtain a similar convergence guarantee on DP-SGD with gradient clipping. {Importantly, we are able to prove such convergence guarantee even \emph{without} Lipschitzness of the loss function, which is often required for DP-SGD analyses.} We also  provide extensive empirical studies to investigate the gradient distributions of DP-SGD across different epoches on two real datasets. To visualize the symmetricity of the gradient distributions, we perform multiple random projections on the gradients and examine the two-dimensional projected distributions. Our results suggest that the gradient distributions in DP-SGD quickly exhibit symmetricity, despite the asymmetricity at initialization.

\textbf{Gradient correction mechanism.} Finally, we provide a simple modification to DP-SGD that can mitigate the clipping bias. We show that perturbing the gradients \emph{before} clipping can provably reduce the clipping bias for any gradient distribution. The pre-clipping perturbation does not by itself provide privacy guarantees, but can trade-off the clipping bias with higher variance.

\subsection{Related work}
The divergence caused by the clipping bias was also studied by prior work. In \citet{pichapati2019adaclip}, an adaptive gradient clipping method is analyzed and the divergence is characterized by a bias depending on the difference between the clipped and unclipped gradients. However, they study a different variant of clipping that bounds the $\ell_\infty$ norm of the gradient instead of $\ell_2$ norm; the latter, which we study in this paper, is the more commonly used clipping operation~\citep{abadi2016deep,tf}.  In \citet{zhang2019gradient}, the divergence is characterized by a bias depending on the clipping probability. These results suggest that, the clipping probability as well as the bias are inversely proportional to the size of the clipping threshold. For example, small clipping threshold results in large bias in the gradient estimation, which can potentially lead to worse training and generalization performance. \citet{om} provides another adaptive gradient clipping heuristic that sets the threshold based on a privately estimated quantile, which can be viewed as minimizing the clipping probability. {In a very recent work, \citet{song2020characterizing} shows that gradient clipping can lead to constant regret in worst case and it is equivalent to huberizing the loss function for generalized linear problems.  Compared with the aforementioned works, our result shows that the bias caused by gradient clipping can be zero when the gradient distribution is symmetric, revealing the effect of gradient distribution beyond clipping probabilities.}

\section{Convergence of SGD with clipped gradient}
\label{sgd}

In this section, we analyze convergence of SGD with clipped gradient, but without the Gaussian perturbation. This simplification is useful for isolating the clipping bias. Consider the standard stochastic optimization formulation 
\begin{align}
    \min_{x} f(x) \triangleq \mathbb E_{s \sim D} [f(x,s)]
\end{align}
where $D$ denotes the underlying distribution over the examples $s$. In the next section, we will instantiate $D$ as the empirical distribution over the private dataset.
We assume that the algorithm is given access to a stochastic gradient oracle: given any iterate $x_t$ of SGD, the oracle returns $\nabla f(x_t) + \xi_t$, where $\xi_t$ is independent noise with zero mean. {In addition, we assume $f(x)$ is G-smooth, i.e. $\|\nabla f(x) - \nabla f(y)\|\leq G \|x - y\|, \forall x,y$. }At each iteration $t$, SGD with gradient clipping performs the following update:
\begin{align}\label{eq:clipping} 
x_{t+1} = x_t - \alpha \clip({\nabla f(x_t) + \xi_t},c): = x_t - \alpha g_t.
\end{align}
where $g_t =  \clip({\nabla f(x_t) + \xi_t},c)$ denotes the realized clipped gradient. 

{To carry out the analysis of iteration \eqref{eq:clipping}, we first note that the standard convergence analysis for SGD-type method consists of two main steps: 

\noindent{\bf S1)} Show that the following key term diminishes to zero: $\mathbb E[\langle \nabla f(x_t), g_t \rangle]$.

\noindent{\bf S2)} Show that the aforementioned quantity is proportional to {$\|\nabla f(x_t)\|^2$ or $c\|\nabla f(x_t)\|$}, indicating that the size of gradient also decreases to zero.

In our analysis below, we will see that showing the first step is relatively easy, while the main challenge is to show that the second step holds true.Our first result is given below.
}

\begin{theorem}\label{thm: general_conv}
Let $G$ be the Lipschitz constant of $\nabla f$ such that $\|\nabla f(x) - \nabla f(y)\|\leq G \|x - y\|, \forall x,y$. For SGD with gradient clipping of threshold $c$, if we set $\alpha = \frac{1}{\sqrt{T}}$, we have
\begin{align}
    \frac{1}{T} \sum_{t=1}^T \mathbb E \left[ \langle \nabla f(x_t), g_t \rangle\right] \leq \frac{D_f}{\sqrt{T} } + \frac{G}{2\sqrt{T}} c^2
\end{align}
where $D_f = f(x_1) - \min_{x}f(x)$.
\end{theorem}
{Note that for SGD without clipping, we have $\mathbb E[\langle \nabla f(x_t), g_t \rangle ] =\|\nabla f(x_t)\|^2$, so the convergence can be readily established.
However, when clipping is applied, the expectation is different but if we have $\mathbb E[\langle \nabla f(x_t), g_t \rangle ]$ being  positive or scaling with $\|\nabla f(x_t)\|$, we can still establish a convergence guarantee. However, the divergence examples (Example 1 and 2) indicate proving this second step requires additional conditions. Now we study a geometric condition that is observed empirically.}

\subsection{Symmetricity-Based Analysis on Gradient Distribution}
Let $p_t(\xi_t)$ be the probability density function of $\xi_t$ and $\tilde{p}_t(\xi_t)$ is an arbitrary distribution. 
 To quantify the clipping bias, we start the analysis with the following decomposition:
{\small
\begin{align}\label{eq: split_bias}
    \mathbb E_{\xi_t \sim p}[\langle \nabla f(x_t), g_t \rangle ] 
    =& \mathbb E_{\xi_t \sim \tilde{p}} [\langle \nabla f(x_t),g_t \rangle] + \underbrace{\int \langle \nabla f(x_t),\clip(\nabla f(x_t)+ \xi_t,c) \rangle (p_t(\xi_t)-\tilde{p_t}( \xi_t)) d\xi_t }_{b_t}
\end{align}
}%
Recall that $\xi_t \sim p$ in \eqref{eq: split_bias} is the gradient noise caused by data sampling, we can choose $\tilde{p}(\xi_t)$ to be some "nice" distribution that can effectively relate $\mathbb E_{\xi_t \sim \tilde{p}} [\langle \nabla f(x_t),g_t \rangle]$ to {$\|\nabla f(x_t)\|^2$} and the remaining term will be treated as the bias. This way of splitting ensures that when the gradients follow a "nice" distribution, the bias will diminish with the distance between $p$ and $\tilde p_t$.
More precisely, we want to find a distribution $\tilde{p}$ such that $\mathbb E_{\xi_t \sim \tilde{p}} [\langle \nabla f(x_t),g_t \rangle]$ is lower bounded by norm squared of the true gradient and thus convergence can be ensured.

{{A straightforward "nice" distribution will be one that can ensure} $\langle \nabla f(x_t),g_t \rangle \geq \Omega(\|\nabla f(x_t)\|_2^2),\ \forall g_t$, i.e. all stochastic gradients are positively aligned with the true gradient.} This may be satisfied when the gradient is large and the noise $\xi$ is bounded.  However, when the gradient is small, it is hard to argue that this can still be true {in general}. {Specifically}, in the training of neural nets, the cosine similarities between many stochastic gradients and the true gradient {(i.e. $\cos (\nabla f(x_t), \nabla f(x_t) + \xi_t$))} can be negative, which implies that this assumption does not hold (see Figure \ref{fig:mnist_cosine} in Section~\ref{sec:experiments}).


Although Figure \ref{fig:mnist_cosine} seems to exclude the {\it ideal} distribution, we observe that the distribution of cosine of the gradients appears to be {\it symmetric}. {Will such a "symmetricity" property help define the "nice" distribution for gradient clipping? If so, how to  characterizes the performance of gradient clipping in this situation? In the following result, we rigorously answer to these questions. }

\begin{theorem}\label{thm: symmetric_descent}
Assume $\tilde p(\xi_t) = \tilde p(-\xi_t)$, gradient clipping with threshold $c$ has the following properties. 
\begin{align*}
\textrm{1. If } \|\nabla f(x_t)\| \leq \frac{3}{4}c,&\quad \text{ then }\quad
\mathbb E_{\xi_t \sim \tilde{p}}[\langle \nabla f(x_t), g_t \rangle ]  \geq \| \nabla f(x_t)\|^2  \mathbb P_{\xi_t \sim \tilde p}\left(\|\xi_t\|<\frac{c}{4}\right)\\
\textrm{2. If } \|\nabla f(x_t)\| > \frac{3}{4}c,&\quad \text{ then } \quad \mathbb E_{\xi_t \sim \tilde{p}}[\langle \nabla f(x_t), g_t \rangle ]  \geq \frac{3}{4}  c \| \nabla f(x_t)\|  \mathbb P_{\xi_t \sim \tilde p}\left(\|\xi_t\|<\frac{c}{4} \right)
\end{align*}
\end{theorem}
Theorem \ref{thm: symmetric_descent} states that when the noise distribution is symmetric, gradient clipping will keep the expected clipped gradients positively aligned with the true gradient. This is the desired property that can guarantee convergence. The probability term characterizes the possible slow down caused by gradient clipping, we provide more discussions and experiments on this term in the Appendix. 

{Combining Theorem \ref{thm: symmetric_descent} with Theorem \ref{thm: general_conv}, we have Corollary \ref{corl: conv} to fully characterize the convergence behavior of SGD with gradient clipping.
\begin{corollary} \label{corl: conv}
Consider the SGD algorithm with gradient clipping given in \eqref{eq:clipping}. 
Set $\alpha = \frac{1}{\sqrt{T}}$, and choose $\tilde p_t(\xi) = \tilde p_t(-\xi)$. Then  the following holds:
\begin{align} \label{eq: pure_conv}
    &\frac{1}{T} \sum_{t=1}^T \mathbb P_{\xi_t \sim \tilde p_t}\left(\|\xi_t\|<\frac{c}{4}\right) \min\left\{\| \nabla f(x_t)\|, \frac{3}{4}c\right\}  \| \nabla f(x_t)\| \leq \frac{D_f}{\sqrt{T} } + \frac{G}{2\sqrt{T}} c^2  - \frac{1}{T}\sum_{t=1}^T b_t,
\end{align}
where we have defined 
$b_t := 
\int \langle \nabla f(x_t),\clip(\nabla f(x_t)+ \xi_t,c) \rangle (p_t(\xi_t)-\tilde p_t(\xi_t)) d\xi_t$.
\end{corollary}

{Therefore, as long as  the probabilities $ \mathbb P_{\xi_t \sim \tilde p_t}\left(\|\xi_t\|<\frac{c}{4}\right)$ are bounded away from 0 and the symmetric distributions $\tilde p_t$ are close approximations to $p_t$ (small bias $b_t$),\footnote{Both Theorem~\ref{thm: symmetric_descent} and Corollary~\ref{corl: conv} hold under a more relaxed condition of $\tilde p(\xi) = \tilde p(-\xi)$ for $\xi$ with $\ell_2$ norm exceeding $c/4$.} then gradient norm goes to 0. 
Moreover, when $\|\xi_t\|$ is drawn from a sub-gaussian distribution with constant variance, the probability does not diminish with the dimension. This is consistent with the observations in recent work of \citet{LiGZCB20, tiny} on deep learning training, and we also provide our own empirical evaluation on the probability term in the Appendix.}

{Note if the bias 
is negative and very large, the bound on the rhs will not be meaningful. 
Therefore, it is useful to further study properties of such bias term.}  In the next section, we will discuss how large the bias term can be for a few choices of $p$ and $\tilde{p}$. {It turns out that the accumulation of $b_t$ can be negative and can help in some cases.}}

\subsection{Beyond symmetric distributions}\label{sec: beyond_symmetric}
 Theorem \ref{thm: symmetric_descent} and Corollary \ref{corl: conv} suggest that as long as the distribution $p$ is sufficiently close to a symmetric distribution $\tilde p$, the convergence bias expressed as $\sum_{t=1}^T b_t$ will be small. {We now show that our bias decomposition result enables us to analyze the effect of the bias even for some highly asymmetric distributions. Note that when $b_t \geq 0$, the bias in fact helps convergence according Corollary \ref{corl: conv}. }
We now provide three examples where $b_t$ can be non-negative. {Therefore, near-symmetricity is not a necessary condition for convergence,  our symmetricity-based analysis remains an effective tool to establish convergence for a broad class of distributions. }

\textbf{Positively skewed.} Suppose $p$ is positively skewed, that is, $p(\xi) \geq p(-\xi)$, for all $\xi$ with $\langle \xi, \nabla f(x) \rangle >0 $. With such distributions, the stochastic gradients tend to be positively aligned with the true gradient. If one chooses $\tilde p (\xi_t) = \frac{1}{2} (p(\xi_t)+p(-\xi_t))$, the bias $b_t$ can be written as  \begin{align*}
    \int_{\xi_t \in \{\xi: \langle \xi, \nabla f(x_t) \rangle > 0\} } \langle \nabla f(x_t),\clip(\nabla f(x_t)+ \xi_t,c) - \clip(\nabla f(x_t)- \xi_t,c) \rangle (\frac{1}{2} (p(\xi_t)-p(-\xi_t))) d\xi_t,
\end{align*}  
which is always positive since $\langle \nabla f(x_t),\clip(\nabla f(x_t)+ \xi_t,c) - \clip(\nabla f(x_t)- \xi_t,c) \rangle \geq 0$. Substituting into \eqref{eq: split_bias}, we have $\mathbb E_{\xi_t \sim p}[\langle \nabla f(x_t), g_t \rangle ]$ strictly larger than $\mathbb E_{\xi_t \sim \tilde{p}} [\langle \nabla f(x_t),g_t \rangle]$, which means the  positive skewness help convergence (we want $\mathbb E_{\xi_t \sim p}[\langle \nabla f(x_t), g_t \rangle ]$ as large as possible).
    
\textbf{Mixture of symmetric.} The distribution of stochastic gradient $\nabla f(x_t) + \xi_t$ is a mixture of two symmetric distributions $p_0$ and $p_1$ with mean $0$ and $v$ respectively. Such a distribution might be possible when most of samples are well classified.  In this case, even though the distribution of $\xi_t$ is not symmetric, one can apply similar argument of Theorem \ref{thm: symmetric_descent} to the component with mean $v$, and the zero mean component yield a bias 0. In particular, let $w_0$ be the probability that $\nabla f(x_t) + \xi_t$ is drawn from $p_0$. One can choose $\tilde p = p - w_0 p_0$ which is  the component symmetric over $v$. The bias become
    \begin{align}
    \int \langle \nabla f(x_t),\clip(\nabla f(x_t)+ \xi_t,c) \rangle w_0 p_0(\xi_t) d\xi_t = 0
\end{align}  
since $p_0(\xi_t)$ corresponds to a zero mean symmetric distribution of $\nabla f(x_t)+ \xi_t$. Note that despite $\tilde p = p - w_0 p_0$ is not a distribution since $\int \tilde p(\xi_t) = 1-w_0$, Theorem \ref{thm: symmetric_descent} can still be applied with everything on RHS of inequalities multiplied by  $1-w_0$ because one can apply Theorem \ref{thm: symmetric_descent} to distribution $\tilde p(\xi_t)/(1-w_0)$ and then scale everything down.

\textbf{Mixture of symmetric or positively skewed.} If $p$ is a mixture of multiple  symmetric or positively skewed distributions, one can split the distributions into multiple ones and use their individual properties. E.g. one can easily establish convergence guarantee for $p$ being a mixture of $m$ {spherical}  distributions with mean $u_1,...,u_m$ and $\langle f(x_t), u_i\rangle \geq 0, \forall i \in [m]$ as in the following theorem. 
\begin{theorem}\label{thm: multimodal}
Given $m$ distributions with the pdf of the $i$th distribution being  $p_i(\xi) = \phi_i(\|\xi - u_i\|)$ for some function $\phi_i$. If $\nabla f(x_t) = \sum_{i=1}^n w_iu_i$ for some $w_i \geq 0,  \sum_{i=1}^m w_i = 1$. Let $p'(\xi) = \sum_{i=1}^m w_i p_i(\xi - \nabla f(x_t) ), $ be a mixture of these distributions with  zero mean. If $\langle u_i, \nabla f(x_t) \rangle \geq 0, \forall i \in [m]$, we have 
\begin{align*}
\mathbb E_{\xi_t \sim {p'}}[\langle \nabla f(x_t), g_t \rangle ]  \geq \| \nabla f(x_t)\| \sum_{i=1}^m w_i\min(\|u_i\|,\frac{3}{4}c)\cos (\nabla f(x_t),u_i)  \mathbb P_{\xi_t \sim p_i}(\|\xi_t\|<\frac{c}{4}) \geq 0
\end{align*}
\end{theorem}

Besides these examples of favorable biases above, there are also many cases where $b_t$ can be negative and lead to a convergence gap, such as negatively skewed distributions or multimodal  distributions with highly imbalanced modes. We have illustrated possible distributions in our divergence examples (Examples 1 and 2). In such cases, one should expect that clipping has an adversarial impact on the convergence guarantee. {However, as we also show in Section \ref{sec:experiments}, the gradient distributions on real datasets tend to be symmetric, and so their clipping bias to be small.}

\section{DP-SGD with Gradient Clipping}

We now extend the results above to  analyze the overall convergence DP-SGD with gradient clipping. To match up with the setting in Section~\ref{sgd}, we consider the distribution $D$ to be the empirical distribution over a private dataset $S$ of $n$ examples $\{s_1, \ldots , s_n\}$, and so $f(x) = \frac{1}{n} \sum_{i=1}^n f(x, s_i)$. For any iterate $x_t\in \mathbb{R}^d$ and example $s_i$, let $\xi_{t, i} = \nabla f(x_t, s_i) - \nabla f(x_t)$ denote the gradient noise on the example, and $p_t$ denote the distribution over $\xi_{t,i}$. At each iteration $t$, DP-SGD performs:
\begin{align} \label{eq: dpsgd}
x_{t+1} = x_t - \alpha \left( \left( \frac{1}{|S_t|} \sum_{i\in S_t} \clip(\nabla f(x_t) + \xi_{t,i},c) \right) + Z_t\right)
\end{align}
where $S_t$ is a random subsample of $S$ (with replacement)\footnote{Alternatively, subsampling with replacement~\citep{WangBK19} and Poisson subsampling~\citep{ZhuW19} have also been proposed.} and $Z_t \sim \mathcal N(0, \sigma^2  I)$ is the noise added for privacy. We first recall the privacy guarantee of the algorithm below:

\begin{theorem}[Privacy (Theorem 1 in \citet{abadi2016deep})]
There exist constants $u$ and $v$ so that given the number of iterations $T$, for any $\epsilon \leq u q^2 T$, where $q = \frac{|S_t|}{n}$,  DP-SGD with gradient clipping of threshold $c$ is $(\epsilon, \delta)$-differentially private for any $\delta >0$, if $\sigma^{2} \geq v \frac{ c^{2} T \ln \left(\frac{1}{\delta}\right)}{n^{2} \epsilon^{2}}$.
\end{theorem}

By accounting for the sub-sampling noise and Gaussian perturbation in DP-SGD, we obtain the following convergence guarantee, where we further bound the clipping bias term $b_t$ with the Wasserstein distance between the gradient distribution and a coupling symmetric distribution.

\begin{theorem}[Convergence]\label{thm: dp_conv}
Let $d$ be the dimensionality of the parameters. For DP-SGD with gradient clipping, if we set  $\alpha = \frac{ \sqrt{D_fd\ln(\frac{1}{\delta})}
}{n\epsilon c\sqrt{L} }$, $\tilde{p}_t(\xi_t) = \tilde{p}_t(-\xi_t)$, {let $m = |S_t|$, there exist $u$ and $v$ such that for any $\epsilon \leq u \frac{m^2}{n^2}T$, $\sigma^{2} = v \frac{ c^{2} T \ln \left(\frac{1}{\delta}\right)}{n^{2} \epsilon^{2}}$},
 we have
\begin{align*}
    \frac{1}{T}\sum_{t=1}^T  \mathbb P_{\xi_t \sim \tilde p} \left(\|\xi_t\|<\frac{c}{4} \right)  h_c(\|\nabla f(x_t)\|)
    \leq & 
    \left(\frac{1}{2 }v+\frac{3}{2}\right) {c}\frac{ \sqrt{D_fGd\ln(\frac{1}{\delta})} 
}{n\epsilon } + \frac{1}{T}\sum_{t=1}^T W_{\nabla f(x_t),c}(\tilde p_t, p_t)
\end{align*}
where $h_c(y)  = \min(y^2, \frac{3}{4}cy)$ and $ W_{v,c}(p,p')$ is the Wasserstein distance between $p$ and $p'$ with metric function $d_{v,c}(a,b) = |\langle v, \clip(v+ a,c) \rangle - \langle v, \clip(v+ b,c) \rangle|$ and $D_f \geq f(x_1) - \min_{x}f(x)$.
\end{theorem}

{\textbf{Remark on convergence rate.}  DP-SGD achieves convergence rate of $O({\sqrt{d}}/{(n\epsilon)})$ in the existing literature. As shown in Theorem \ref{thm: dp_conv}, with gradient clipping, the rate becomes $O({\sqrt{d}}/{(n\epsilon)}) + \text{clipping  bias}$. When gradient distribution is symmetric, the convergence rate of $O({\sqrt{d}}/{(n\epsilon)})$ can be recovered. This result implies that when gradient distribution is symmetric, the clipping operation will only affect the convergence rate of DP-SGD by a constant factor. In addition, since the clipping bias is the Wasserstein distance between the empirical gradient distribution and an oracle symmetric distribution, it can be small when the gradient distribution is approximately symmetric.}

{\textbf{Remark on the Wasserstein distance.} In \eqref{eq: pure_conv}, it is clear that the convergence bias $b_t$ can be bounded by the total variation distance between $p_t$ and $\tilde p_t$. However, this bound becomes trivial when $p_t$ is the empirical distribution over a finite sample, {because} {the total variation distance is always 2 when $\tilde p$ is continuous}. In addition, the bias is hard to interpret when without further transformation. }
This is why we bound $b_t$ by the Wasserstein distance as follows:  
{\small
\begin{align}
    -b_t = &
\int \langle \nabla f(x_t),\clip(\nabla f(x_t)+ \xi_t,c) \rangle (\tilde p(\xi_t)-{p}(\xi_t)) d\xi_t \nonumber \\
=& \int \langle \nabla f(x_t),\clip(\nabla f(x_t)+ \xi_t,c) \rangle \tilde p(\xi_t) d\xi_t - \int \langle \nabla f(x_t),\clip(\nabla f(x_t)+ \xi_t',c) \rangle  p(\xi_t') d\xi_t'  \nonumber \\
=& \int \int (\langle \nabla f(x_t),\clip(\nabla f(x_t)+ \xi_t,c) \rangle - \langle \nabla f(x_t), \clip(\nabla f(x_t)+ \xi_t',c) \rangle) \gamma (\xi_t,\xi_t')  d\xi_t d\xi'_t \nonumber \\
\leq & \int \int |\langle \nabla f(x_t),\clip(\nabla f(x_t)+ \xi_t,c) \rangle - \langle \nabla f(x_t), \clip(\nabla f(x_t)+ \xi_t',c) \rangle| \gamma (\xi_t,\xi_t')  d\xi_t d\xi'_t
\end{align}
}%
where $\gamma$ is any joint distribution with marginal $\tilde p$ and $ p$. Thus, we have 
\begin{align}
    b_t \leq \inf_{\gamma \in \Gamma(\tilde p, p)}\int \int |\langle \nabla f(x_t),\clip(\nabla f(x_t)+ \xi_t,c) \rangle - \langle \nabla f(x_t), \clip(\nabla f(x_t)+ \xi_t',c) \rangle| \gamma (\xi_t,\xi_t')  d\xi_t d\xi'_t \nonumber
\end{align}
where $\Gamma(\tilde p,  p)$ is the set of all couplings with marginals $\tilde p$ and $ p$ on the two factors, respectively. If we define the distance function $d_{y,c}(a,b) = |\langle y, \clip(y+ a,c) \rangle - \langle y, \clip(y+ b,c) \rangle|$, we have 
{\small
\begin{align}
    b_t \leq \inf_{\gamma \in \Gamma(\tilde p, p)}\int \int d_{\nabla f(x_t), c}(\xi_t, \xi_t') \gamma (\xi_t,\xi_t')  d\xi_t d\xi'_t 
\end{align}
}%
which is Wasserstein distance defined on the distance function $d_{\nabla f(x_t), c}$ {and it converges to the distance between the population distribution of gradient and $\tilde p$ with $n$ being large. Thus, if the population distribution of gradient is approximate symmetric, the bias term tends to be small.} In addition, the distance function is  uniformly bounded by $\|\nabla f(x)\| c$ {which makes it is more favorable than $\ell_2$ distance}. Compared with the expression of $b_t$ in Corollary \ref{corl: conv}, the Wasserstein distance is easier to interpret when $\tilde p$ is discrete.

\section{Experiments}\label{sec:experiments}
In this section, we investigate whether the gradient distributions of DP-SGD are approximate symmetric in practice. However, since the gradient distributions are high-dimensional, certifying symmetricity is in general intractable. We instead consider two simple proxy measures and visualizations. 

\begin{figure}[htbp]
	\centering
	\begin{minipage}[c]{0.24\linewidth}
		\includegraphics[width=\linewidth]{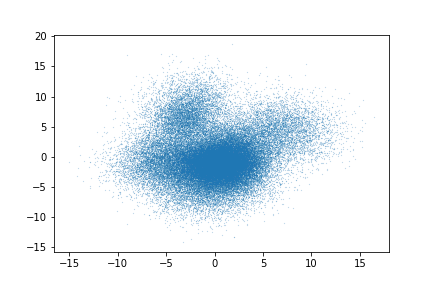}
	\end{minipage}
	\begin{minipage}[c]{0.24\linewidth}
		\includegraphics[width=\linewidth]{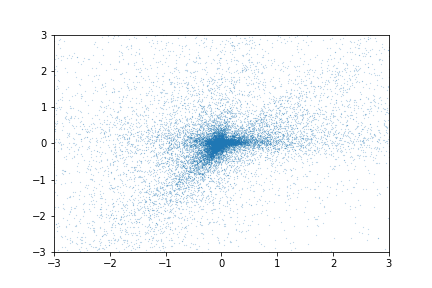}
	\end{minipage}
	\begin{minipage}[c]{0.24\linewidth}
		\includegraphics[width=\linewidth]{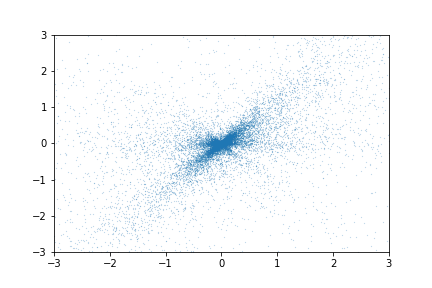}
	\end{minipage}
	\begin{minipage}[c]{0.24\linewidth}
		\includegraphics[width=\linewidth]{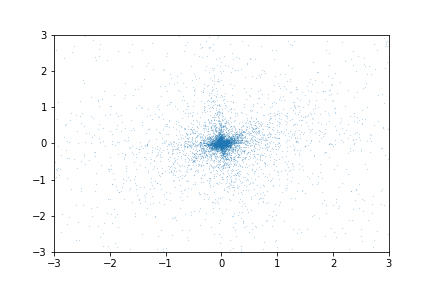}
	\end{minipage}

	\begin{minipage}[c]{0.24\linewidth}
		\includegraphics[width=\linewidth]{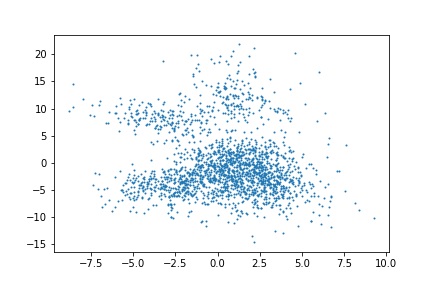}
		\centering{\small (a) Epoch 0}
	\end{minipage}
	\begin{minipage}[c]{0.24\linewidth}
		\includegraphics[width=\linewidth]{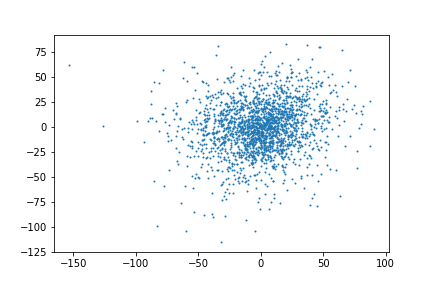}
		\centering{\small (b) Epoch 5}
	\end{minipage}
	\begin{minipage}[c]{0.24\linewidth}
		\includegraphics[width=\linewidth]{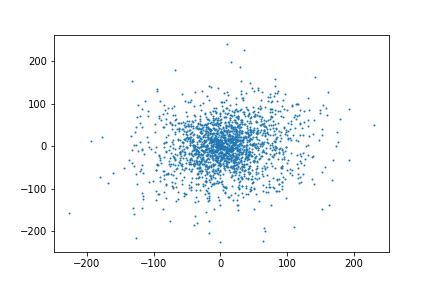}
		\centering{\small (c) Epoch 10}
	\end{minipage}
	\begin{minipage}[c]{0.24\linewidth}
		\includegraphics[width=\linewidth]{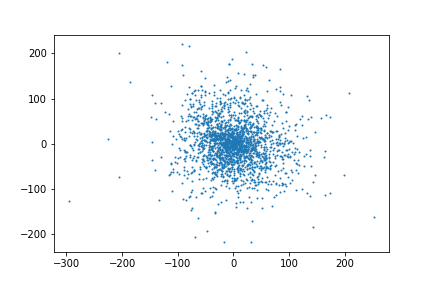}
		\centering{\small (d) Epoch 59}
	\end{minipage}
	\caption{\small Gradient distributions on MNIST (top row) and CIFAR-10 (bottom row)  at the end of different epochs (indexed by columns). The gradients for epoch 0 are computed at initialization (before training). 
	\vspace{-0.4cm}
	}
    \label{fig:cifar}
\end{figure}

\xhdr{Setup.} We run DP-SGD implemented in Tensorflow \footnote{https://github.com/tensorflow/privacy/tree/master/tutorials} on two popular datasets MNIST \citep{lecun2010mnist} and CIFAR-10 \citep{krizhevsky2009learning}. For MNIST, we train a CNN with two convolution layers with 16 4$\times$4 kernels followed by a fully connected layer with 32 nodes. We use DP-SGD to train the model with $\alpha = 0.15$, and a batchsize of 128. For CIFAR-10, we train a CNN with two convolutional layers with 2$\times$2 max pooling of stride 2 followed by a fully connected layer, all 
using ReLU activation, each layer uses a dropout rate of 0.5. The two convolution layer has 32 and 64 3$\times$3 kernels, the fully connected layer has 1500 nodes. We use $\alpha = 0.001$ and decrease it by 10 times every 20 epochs.  The clip norm of both experiments is set to be $c=1$ and the noise multiplier is 1.1.

\begin{figure}[htbp]
	\centering
	\begin{minipage}[c]{0.24\linewidth}
		\includegraphics[width=\linewidth]{figures/mnist/epoch9/mnist_gradient_9_0.png}
		\centering{\small (a) Repeat 1}
	\end{minipage}
	\begin{minipage}[c]{0.24\linewidth}
		\includegraphics[width=\linewidth]{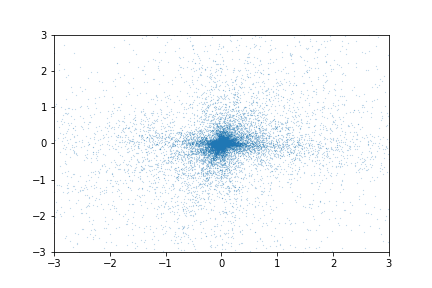}
		\centering{\small (b) Repeat 2}
	\end{minipage}
	\begin{minipage}[c]{0.24\linewidth}
		\includegraphics[width=\linewidth]{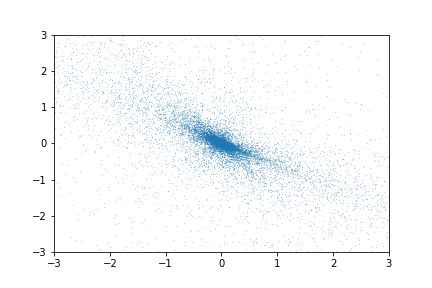}
		\centering{\small (c) Repeat 3}
	\end{minipage}
	\begin{minipage}[c]{0.24\linewidth}
		\includegraphics[width=\linewidth]{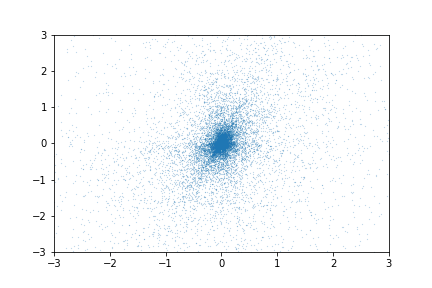}
		\centering{\small (d) Repeat 4}
	\end{minipage}
	\\ 
		\begin{minipage}[c]{0.24\linewidth}
		\includegraphics[width=\linewidth]{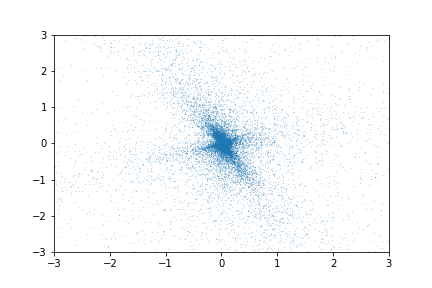}
		\centering{\small (e) Repeat 5}
	\end{minipage}
	\begin{minipage}[c]{0.24\linewidth}
		\includegraphics[width=\linewidth]{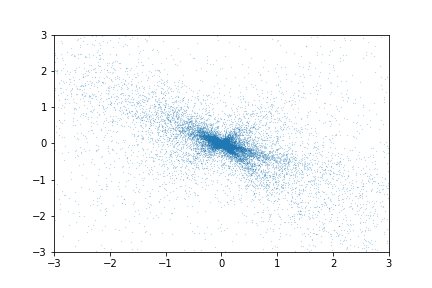}
		\centering{\small (f) Repeat 6}
	\end{minipage}
	\begin{minipage}[c]{0.24\linewidth}
		\includegraphics[width=\linewidth]{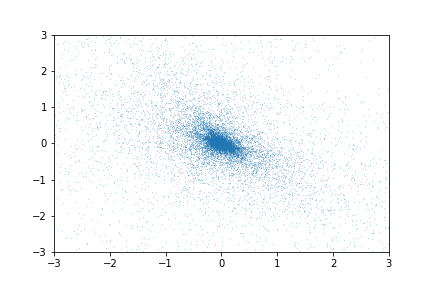}
		\centering{\small (g) Repeat 7}
	\end{minipage}
	\begin{minipage}[c]{0.24\linewidth}
		\includegraphics[width=\linewidth]{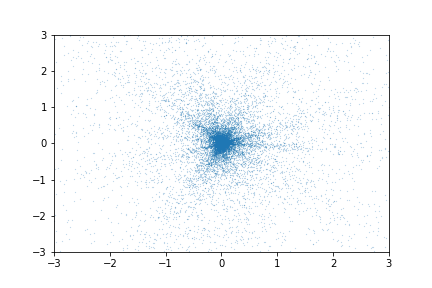}
		\centering{\small (h) Repeat 8}
	\end{minipage}
	\caption{\small Gradient distributions on MNIST at the end of epoch 9 projected using different random matrices. 
	}
    \label{fig:mnist_ep9_main}
\end{figure}

\xhdr{Visualization with random projections.} We visualize the gradient distribution by projecting the gradient to a two-dimensional space using random Gaussian matrices. Note that given any symmetric distribution, its two-dimensional projection remains symmetric for any projection matrix. On the contrary, if for all projection matrix, the projected gradient distribution is symmetric, the original gradient distribution should also be symmetric. We repeat the projection using different randomly generated matrices and visualize the induced distributions.

From Figure \ref{fig:cifar}, we can see that on both datasets, the gradient distribution is non-symmetric before training (Epoch 0), but over the epochs, the gradient distributions become increasingly symmetric. The distribution of gradients  on MNIST at the end of epoch 9 projected to a random two-dimensional space using different random matrices is shown in Figure \ref{fig:mnist_ep9_main}. It can be seen that the approximate symmetric property holds for all 8 realizations. We provide many more visualizations from different realized random projections across different epochs in the Appendix.

\begin{figure}[htbp]
 \vspace{-0.4cm}
	\centering
	\begin{minipage}[c]{0.3\linewidth}
		\includegraphics[width=\linewidth]{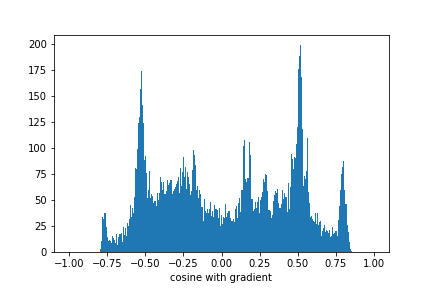}
		\centering{\small (a) Epoch 4}
	\end{minipage}
	\begin{minipage}[c]{0.3\linewidth}
		\includegraphics[width=\linewidth]{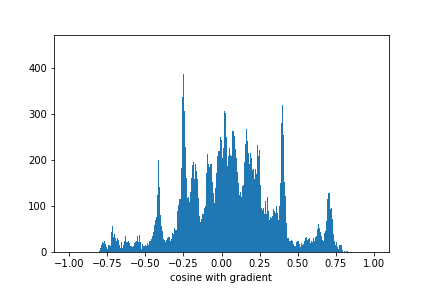}
		\centering{\small (b) Epoch 10}
	\end{minipage}
	\begin{minipage}[c]{0.3\linewidth}
		\includegraphics[width=\linewidth]{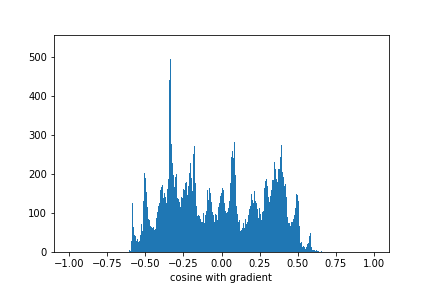}
		\centering{\small (c) Epoch 59}
	\end{minipage}
	\caption{\small  Histogram of cosine between stochastic gradients and the true gradient at the end of different epochs.
	}
    \label{fig:mnist_cosine}
    \vspace{-0.2cm}
\end{figure}

\xhdr{Symmetricity of angles.} We also measure the cosine similarities between per-sample stochastic gradients and the true gradient. We observe that the cosine similarities between per-sample stochastic gradients  and the true gradient  (i.e. $\cos(\nabla f(x_t)+ \xi_{t,i}, \nabla f(x_t))$)  is approximate symmetric around 0 as shown in the histograms in Figure \ref{fig:mnist_cosine}.

\section{Mitigating Clipping Bias with Perturbation}\label{sec: bias_reduce}
\begin{figure}[htbp]
\vspace{-0.3cm}
	\centering
	\begin{minipage}[c]{0.32\linewidth}
		\includegraphics[width=\linewidth]{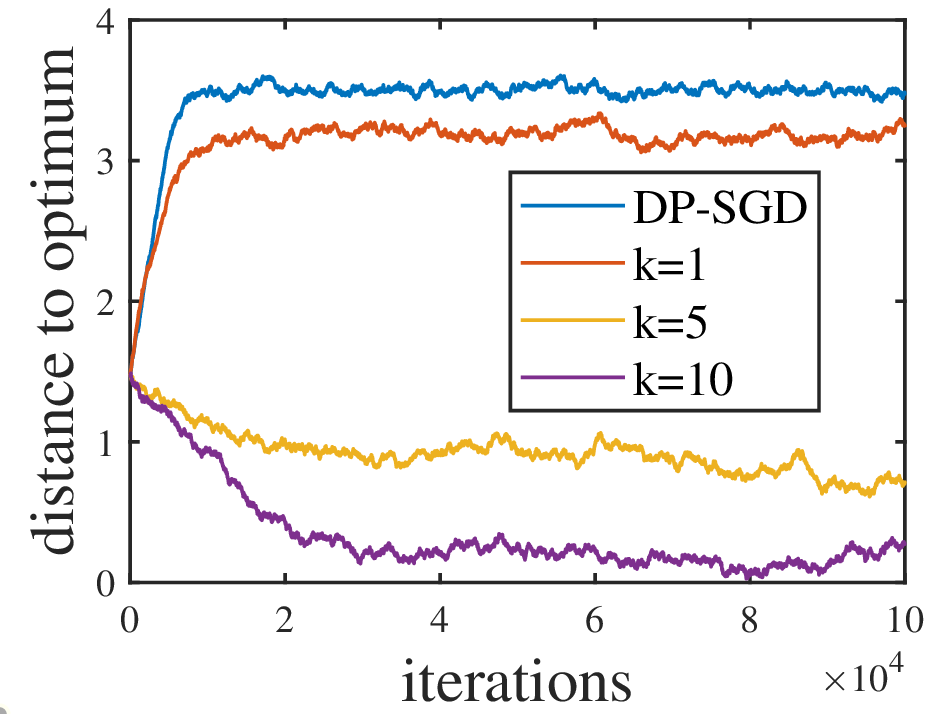}
		\centering{\small (a) Example 1}
	\end{minipage}
	\begin{minipage}[c]{0.32\linewidth}
		\includegraphics[width=\linewidth]{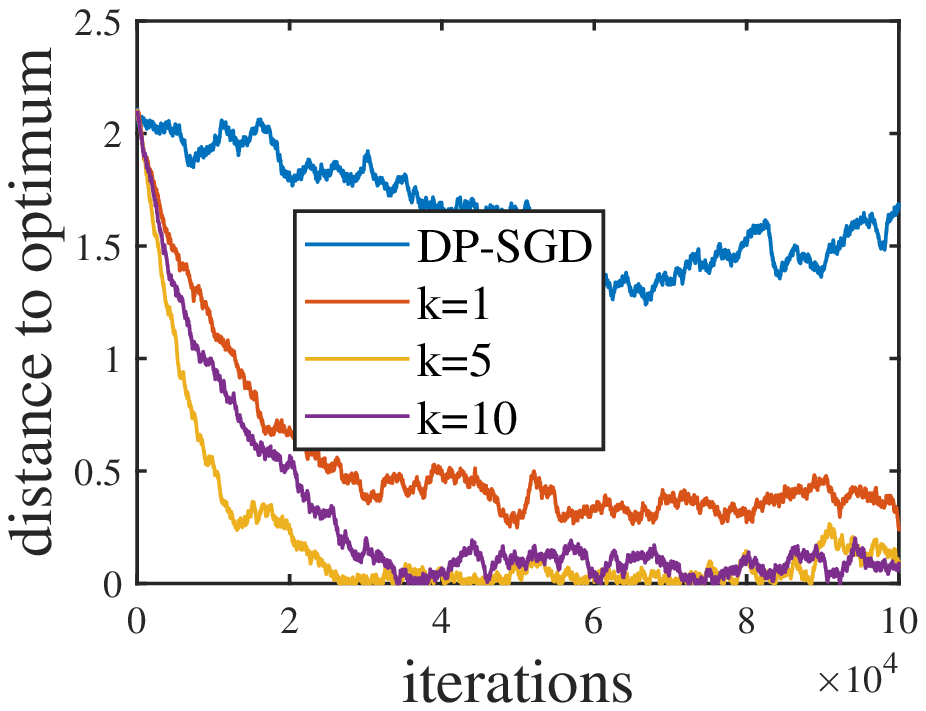}
		\centering{\small (b) Example 2}
	\end{minipage}
		\begin{minipage}[c]{0.32\linewidth}
		\includegraphics[width=\linewidth]{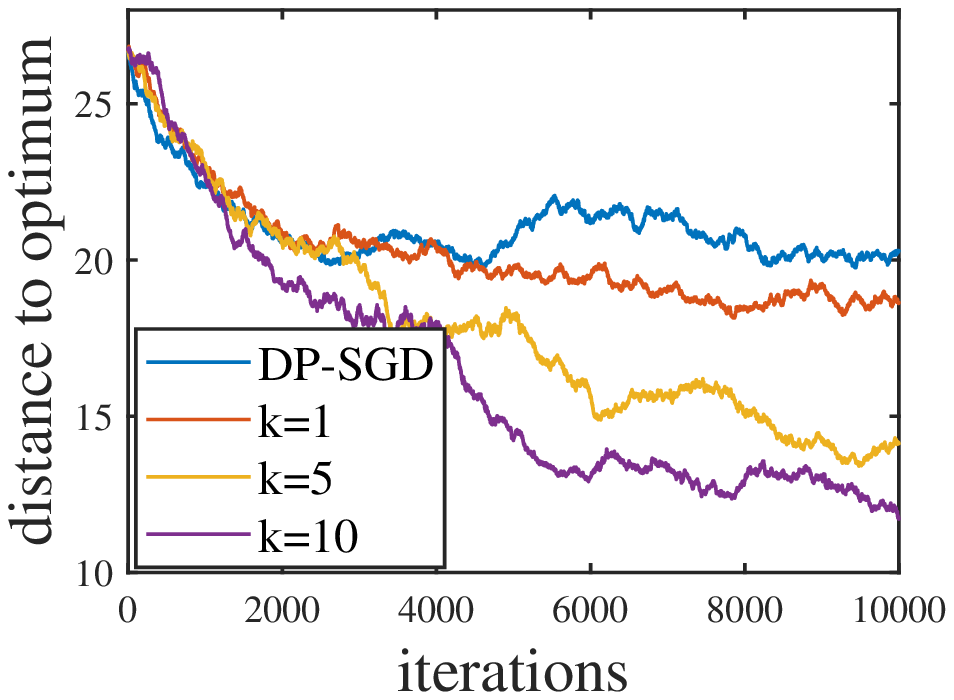}
		\centering{\small (b) Synthetic data}
	\end{minipage}
    \label{fig:example}
        \caption{Effect of pre-clipping perturbation on the examples.}
    \vspace{-0.3cm}
\end{figure}

From previous analyses, SGD with gradient clipping and DP-SGD have good convergence performance when the gradient noise distribution is approximately symmetric or when the gradient bias favors convergence (e.g., mixture of symmetric distributions with aligned mean). {Although in practice, gradient distributions do exhibit (approximate) symmetry (see Sec. \ref{sec:experiments}), it would be desirable to have tools to handle situations where the clipping bias does not favor convergence.} {Now we provide an approach to decrease the bias. If one adds some Gaussian noise before clipping, i.e. 
\begin{align}
    g_t = \clip(\nabla f(x_t)+ \xi_t + k \zeta_t,c),  \zeta_t \sim \mathcal N(0,  I),
\end{align}  we can prove $|b_t| = O\left(\frac{\sigma^2_{\xi_t}}{k^2}\right)$ as in Theorem \ref{thm: scale_noise_main}.
\begin{theorem}\label{thm: scale_noise_main}
Let $g_t = \clip(\nabla f(x_t)+ \xi_t + k \zeta_t,c)$ and $\zeta_t \sim \mathcal N(0,  I)$. Then gradient clipping algorithm has following properties: 
\begin{align}
\mathbb E_{\xi_t \sim {p}, \zeta_t}[\langle \nabla f(x_t), g_t \rangle ]  \geq \| \nabla f(x_t)\| \min\left\{\| \nabla f(x_t)\|, \frac{3}{4}c\right\}  \mathbb P(\|k\zeta_t\|<\frac{c}{4}) - O(\frac{\sigma^2_{\xi_t}}{k^2})
\end{align} 
where $\sigma^2_{\xi_t}$ is the variance of the gradient noise $\xi_t$.
\end{theorem}

 More discussion can be found in the Appendix. Note that when the perturbation approach is applied to DP-SGD, the update rule \eqref{eq: dpsgd} becomes
\begin{align}
x_{t+1} = x_t - \alpha \bigg( \bigg( \frac{1}{|S_t|} \sum_{i\in S_t} \clip(\nabla f(x_t) + \xi_{t,i} + k\zeta_{t,i},c) \bigg) + Z_t\bigg), \nonumber
\end{align}
where each per-sample stochastic gradient is be perturbed by the noise. By adding the noise, one trade-offs bias with variance. Larger noise make the algorithm converges possibly slower but better. This trick can be helpful when the gradient distribution is not favorable.} To verify its effect in practice, we run DP-SGD with gradient clipping on a few unfavorable problems including examples in Section \ref{sec:intro} and a new high dimensional example. We set $\sigma = 1$ on all the examples (i.e. $Z_t \sim \mathcal N(0,I)$). For the new example, we minimize the function $f(x) = \frac{1}{n}\sum_{i=1}^n\frac{1}{2}\|x-z_i\|^2$ with $n=10000$. Each $z_i$ is drawn from a mixture of isotropic Gaussian with 3 components of dimension 10. The covariance matrix of all components  is $I$ and the means of the 3 components are drawn from $\mathcal N(0,36 I)$, $\mathcal N(0,4 I)$, $\mathcal N(0, I)$, respectively. We set $\alpha = 0.015$ for the new examples and $\alpha = 0.001$ for the examples in Section \ref{sec:intro}. Figure \ref{fig:example} shows $\|x_t - \argmin_x f(x)\|$ versus $t$. We can see DP-SGD with gradient clipping converges to non-optimal points as predicted by theory. {In contrast, pre-clipping perturbation ensures convergence. }

\section{Conclusion and Future Work}
In this paper, we provide a theoretical analysis on the effect of gradient clipping in SGD and private SGD. We provide a new way to quantify the clipping bias by coupling the gradient distribution with a geometrically symmetric distribution. Combined with our empirical evaluation showing that gradient distribution of private SGD follows some symmetric structure along the trajectory, these results provide an explanation why gradient clipping works in practice. We also provide a perturbation-based technique to reduce the clipping bias even for adversarial instances.

{There are some interesting directions for future work. One main message of this paper is that when gradient distribution is symmetric, gradient clipping will not be detrimental to the performance of DP-SGD. Thus, looking for methods to symmetrify the gradient distribution could be an interesting topic. Another interesting direction is to study gradient distribution of different types of models empirically. We notice the gradient distribution of CNNs on MNIST and CIFAR-10 might be symmetric and a clipping threshold around 1 works well. However, \citet{mcmahan2017learning} found  a relatively large clipping threshold around 10 works best for LSTMs. This implies the gradient distribution on some models might be less symmetric and a concrete empirical analysis on it could motivate future research. Finally, it could be interesting to investigate properties of gradient clipping on a broader class of gradient distributions beyond symmetricity.}

\section*{Acknowledgement}
The research is supported in part by a NSF grant CMMI-1727757, a Google Faculty Research Award, a J.P. Morgan Faculty Award, and a Facebook Research Award.
\bibliography{reference}

\begin{thebibliography}{16}
\providecommand{\natexlab}[1]{#1}
\providecommand{\url}[1]{\texttt{#1}}
\expandafter\ifx\csname urlstyle\endcsname\relax
  \providecommand{\doi}[1]{doi: #1}\else
  \providecommand{\doi}{doi: \begingroup \urlstyle{rm}\Url}\fi

\bibitem[Abadi et~al.(2016{\natexlab{a}})Abadi, Agarwal, Barham, Brevdo, Chen,
  Citro, Corrado, Davis, Dean, Devin, et~al.]{tf}
Mart{\'\i}n Abadi, Ashish Agarwal, Paul Barham, Eugene Brevdo, Zhifeng Chen,
  Craig Citro, Greg~S Corrado, Andy Davis, Jeffrey Dean, Matthieu Devin, et~al.
\newblock Tensorflow: Large-scale machine learning on heterogeneous distributed
  systems.
\newblock \emph{arXiv preprint arXiv:1603.04467}, 2016{\natexlab{a}}.

\bibitem[Abadi et~al.(2016{\natexlab{b}})Abadi, Chu, Goodfellow, McMahan,
  Mironov, Talwar, and Zhang]{abadi2016deep}
Martin Abadi, Andy Chu, Ian Goodfellow, H~Brendan McMahan, Ilya Mironov, Kunal
  Talwar, and Li~Zhang.
\newblock Deep learning with differential privacy.
\newblock In \emph{Proceedings of the 2016 ACM SIGSAC Conference on Computer
  and Communications Security}, pages 308--318, 2016{\natexlab{b}}.

\bibitem[Bassily et~al.(2014)Bassily, Smith, and Thakurta]{basm14}
Raef Bassily, Adam Smith, and Abhradeep Thakurta.
\newblock Private empirical risk minimization: Efficient algorithms and tight
  error bounds.
\newblock In \emph{2014 IEEE 55th Annual Symposium on Foundations of Computer
  Science}, pages 464--473. IEEE, 2014.

\bibitem[Dwork et~al.(2006)Dwork, McSherry, Nissim, and Smith]{DMNS06}
Cynthia Dwork, Frank McSherry, Kobbi Nissim, and Adam Smith.
\newblock Calibrating noise to sensitivity in private data analysis.
\newblock In \emph{Theory of Cryptography Conference}, pages 265--284.
  Springer, 2006.

\bibitem[Gur-Ari et~al.(2018)Gur-Ari, Roberts, and Dyer]{tiny}
Guy Gur-Ari, Daniel~A Roberts, and Ethan Dyer.
\newblock Gradient descent happens in a tiny subspace.
\newblock \emph{arXiv preprint arXiv:1812.04754}, 2018.

\bibitem[Krizhevsky et~al.(2009)Krizhevsky, Hinton,
  et~al.]{krizhevsky2009learning}
Alex Krizhevsky, Geoffrey Hinton, et~al.
\newblock Learning multiple layers of features from tiny images.
\newblock 2009.

\bibitem[LeCun et~al.(2010)LeCun, Cortes, and Burges]{lecun2010mnist}
Yann LeCun, Corinna Cortes, and CJ~Burges.
\newblock Mnist handwritten digit database.
\newblock \emph{ATT Labs [Online]. Available:
  http://yann.lecun.com/exdb/mnist}, 2, 2010.

\bibitem[Li et~al.(2020)Li, Gu, Zhou, Chen, and Banerjee]{LiGZCB20}
Xinyan Li, Qilong Gu, Yingxue Zhou, Tiancong Chen, and Arindam Banerjee.
\newblock Hessian based analysis of sgd for deep nets: Dynamics and
  generalization.
\newblock In \emph{Proceedings of the 2020 SIAM International Conference on
  Data Mining}, pages 190--198. SIAM, 2020.

\bibitem[McMahan et~al.(2017)McMahan, Ramage, Talwar, and
  Zhang]{mcmahan2017learning}
H~Brendan McMahan, Daniel Ramage, Kunal Talwar, and Li~Zhang.
\newblock Learning differentially private recurrent language models.
\newblock \emph{arXiv preprint arXiv:1710.06963}, 2017.

\bibitem[Pichapati et~al.(2019)Pichapati, Suresh, Yu, Reddi, and
  Kumar]{pichapati2019adaclip}
Venkatadheeraj Pichapati, Ananda~Theertha Suresh, Felix~X Yu, Sashank~J Reddi,
  and Sanjiv Kumar.
\newblock Adaclip: Adaptive clipping for private sgd.
\newblock \emph{arXiv preprint arXiv:1908.07643}, 2019.

\bibitem[Song et~al.(2020)Song, Thakkar, and Thakurta]{song2020characterizing}
Shuang Song, Om~Thakkar, and Abhradeep Thakurta.
\newblock Characterizing private clipped gradient descent on convex generalized
  linear problems.
\newblock \emph{arXiv preprint arXiv:2006.06783}, 2020.

\bibitem[Thakkar et~al.(2019)Thakkar, Andrew, and McMahan]{om}
Om~Thakkar, Galen Andrew, and H~Brendan McMahan.
\newblock Differentially private learning with adaptive clipping.
\newblock \emph{arXiv preprint arXiv:1905.03871}, 2019.

\bibitem[Wang and Xu(2019)]{Wang019}
Di~Wang and Jinhui Xu.
\newblock Differentially private empirical risk minimization with smooth
  non-convex loss functions: A non-stationary view.
\newblock In \emph{Proceedings of the AAAI Conference on Artificial
  Intelligence}, volume~33, pages 1182--1189, 2019.

\bibitem[Wang et~al.(2019)Wang, Balle, and Kasiviswanathan]{WangBK19}
Yu-Xiang Wang, Borja Balle, and Shiva~Prasad Kasiviswanathan.
\newblock Subsampled r{\'e}nyi differential privacy and analytical moments
  accountant.
\newblock In \emph{The 22nd International Conference on Artificial Intelligence
  and Statistics}, pages 1226--1235, 2019.

\bibitem[Zhang et~al.(2019)Zhang, He, Sra, and Jadbabaie]{zhang2019gradient}
Jingzhao Zhang, Tianxing He, Suvrit Sra, and Ali Jadbabaie.
\newblock Why gradient clipping accelerates training: A theoretical
  justification for adaptivity.
\newblock In \emph{International Conference on Learning Representations}, 2019.

\bibitem[Zhu and Wang(2019)]{ZhuW19}
Yuqing Zhu and Yu-Xiang Wang.
\newblock Poission subsampled r{\'e}nyi differential privacy.
\newblock In \emph{International Conference on Machine Learning}, pages
  7634--7642, 2019.

\end{thebibliography}
\bibliographystyle{plainnat}

\newpage 
\appendix
 \section{Proof of Theorem \ref{thm: general_conv}}
 By smoothness assumption, we have 
 \begin{align}
    f(x_{t+1}) \leq f(x_t) + \langle \nabla f(x_t), x_{t+1} - x_t \rangle + \frac{1}{2}G \|x_{t+1} - x_t\|^2.
\end{align}
Then, by update rule and the fact that $\|g_t\| \leq c$, we have
 \begin{align}
    f(x_{t+1}) \leq &f(x_t) -\alpha \langle \nabla f(x_t), g_t \rangle + \frac{1}{2}G \alpha^2\|g_t\|^2 \nonumber \\
    \leq& f(x_t) -\alpha \langle \nabla f(x_t), g_t \rangle + \frac{1}{2}G \alpha^2 c^2.
\end{align}
Take expectation, sum over $t$ from 1 to $T$, divide both sides by $T\alpha$, rearranging and substituting into $\alpha = \frac{1}{\sqrt{T}}$, we get
 \begin{align}
   \frac{1}{T} \sum_{t=1}^T \mathbb E [ \langle \nabla f(x_t), g_t \rangle ]   \leq & \frac{1}{T\alpha} ( f(x_1) - f(x_{T+1}) )  + \frac{1}{2}G 
   \alpha c^2 \nonumber \\
   \leq & \frac{1}{\sqrt{T}}\mathbb E [ f(x_1) - f(x_{T+1}) ]  + \frac{1}{2\sqrt{T}} G c^2 \nonumber \\
   \leq & \frac{1}{\sqrt{T}}D_f  + \frac{1}{2\sqrt{T}} G c^2 
\end{align}
where $D_f = f(x_1) - \min_x f(x)$. \hfill $\square$

\section{Proof of Theorem \ref{thm: symmetric_descent}}
In the proof, we assume $\xi_t \sim \tilde p_t$ we omit subscript of $\mathbb P$  and $\mathbb E$ to simplify notations.
\subsection{When gradient is small}
Let us  first consider the case with $\|\nabla f(x_t)\| \leq \frac{3}{4}c$.

Denote $B$ to be the event that $\|\nabla f(x_t) + \xi_t\| \leq c$ and $\|\nabla f(x_t) - \xi_t\| \leq c$, we have $\mathbb P (B) \geq \mathbb P (\|\xi_t\| \leq \frac{c}{4})$. Define $D = \{\xi: \|\nabla f(x_t) + \xi_t\| > c\  \textrm{or}\  \|\nabla f(x_t) - \xi_t\| > c \}$. Taking an expectation conditioning on $x_t$, we have
\begin{align}
    &\mathbb E[\langle \nabla f(x_t), g_t \rangle ] \nonumber \\
    =&  \langle \nabla f(x_t), \mathbb E[ \clip(\nabla f(x_t) + \xi_t, c)] \rangle  \nonumber \\
    =&  \lla \nabla f(x_t), \mathbb E\left[ \clip(\nabla f(x_t) + \xi_t, c)\bigg| B \right]\rra \mathbb P(B) \nonumber \\
    & +  \lla \nabla f(x_t), \mathbb \int_{D  } \clip(\nabla f(x_t) + \xi_t,c) \tilde p(\xi_t) d \xi_t 
    \rra  \nonumber \\
    \geq &  \| \nabla f(x_t)\|^2 \mathbb P(\|\xi_t\|\leq \frac{c}{4}) +  \underbrace{\lla \nabla f(x_t), \mathbb \int_{D} \clip(\nabla f(x_t) + \xi_t,c) \tilde p(\xi_t) d \xi_t 
    \rra}_{T_1}  \nonumber     
\end{align}
where the last inequality is due to $\clip(\nabla f(x_t) + \xi_t, c) = \nabla f(x_t) + \xi_t$ when $B$ happens and $\mathbb P (B) \geq \mathbb P (\|\xi_t\| \leq \frac{c}{4})$ and $\tilde p(\xi) = \tilde p(-\xi)$. 

Now we need to look at $T_1$.

We have 
\begin{align}
    T_1 
    = & \frac{1}{2 }\left(\lla \nabla f(x_t), \mathbb \int_{D} \clip(\nabla f(x_t) + \xi_t,c) \tilde p(\xi_t) d \xi_t
    \rra + \lla \nabla f(x_t), \mathbb \int_{D} \clip(\nabla f(x_t) - \xi_t,c) \tilde p(\xi_t) d \xi_t 
    \rra\right) \nonumber \\
    = & \frac{1}{2 }\lla \nabla f(x_t), \mathbb \int_{D}\left( \clip(\nabla f(x_t) + \xi_t,c) + \clip(\nabla f(x_t) - \xi_t,c)\right) \tilde p(\xi_t) d \xi_t 
    \rra \nonumber \\
     = & \frac{1}{2 } \|\nabla f(x_t)\| \nonumber \\
     &\times  \int_{D}   \underbrace{\left( \|  \clip(\bar g_t + \xi_t,c)\| cos(\bar g_t, \bar g_t + \xi_t)+ \| \clip(\bar g_t - \xi_t,c)\| cos(\bar g_t, \bar g_t- \xi_t) \right)}_{T_2(\xi_t)} \tilde p(\xi_t) d \xi_t
\end{align}
where $ \bar g_t \triangleq \nabla f(x_t)$ and the last equality is because $\langle a,b \rangle = \|a\| \|b\| \cos(a,b)$ for any vector $a,b$, {and that the clipping operation keeps directions}.

Now it reduces to analyzing $T_2(\xi_t)$. Our target now is to prove  $T_2(\xi_t) \geq 0 $.

Let us first consider the case where $||\bar g_t+\xi_t|| \geq c $ and $||\bar g_t-\xi_t|| \geq c $. In this case, we have
\begin{align}
    T_2(\xi) =  c( cos(\bar g_t, \bar g_t + \xi_t)+  cos(\bar g_t, \bar g_t- \xi_t) ) \geq 0 
\end{align}
where the inequality is due to Lemma \ref{lem: cos}.

Another case is one of $||\bar g_t+\xi_t||$ and  $||\bar g_t-\xi_t||$ is less than $c$. Assume $ \cos(\bar g_t, \bar g_t- \xi_t) <0$.  In this case, we must have $\cos(\bar g_t,- \xi_t) <0$ and $ cos(\bar g_t,\bar g_t + \xi_t) >0$ from basic properties of trigonometric functions. Then,  from Lemma \ref{lem: norm_to_cos}, we have
\begin{align}
    ||\bar g_t+\xi|| \geq ||\bar g_t-\xi||.
\end{align} 
So in this case, we have
\begin{align}
   T_2(\xi_t) =&   \|  \clip(\bar g_t + \xi_t,c)\| \cos(\bar g_t, \bar g_t + \xi_t)+ \| \clip(\bar g_t - \xi,c)\| cos(\bar g_t, \bar g_t- \xi) \nonumber \\
    = & c \cdot \cos(\bar g_t, \bar g_t + \xi_t)+ \| \clip(\bar g_t - \xi_t,c)\| \cos(\bar g_t, \bar g_t- \xi_t) \nonumber \\
    \geq  & c \cdot \cos(\bar g_t, \bar g_t + \xi_t)+ c \cdot cos(\bar g_t, \bar g_t- \xi_t) \nonumber \\
    \geq & 0
\end{align}
where the last inequality is due to Lemma \ref{lem: cos}.

Similar argument applies to the case with $ cos(\bar g_t, \bar g_t + \xi_t) <0$.

For the case with $ cos(\bar g_t, \bar g_t+ \xi_t) \geq 0$ and $ cos(\bar g_t, \bar g_t- \xi_t) \geq 0$, we trivially have $T_2(\xi_t) \geq 0$. Thus, we have
\begin{align}
    &\mathbb E[\langle \nabla f(x_t),  g_t \rangle ] 
    \geq   \| \nabla f(x_t)\|^2 \mathbb P(\|\xi_t\|\leq \frac{c}{4}).    
\end{align}
This completes the proof. \qed

\subsection{When gradient is large}
Now let us look at the case where gradient is large, i.e. $\|\nabla f(x_t)\| \geq \frac{3}{4}c$. 

By definition, we have
\begin{align}\label{eq: T_7}
    &\mathbb E[\langle \nabla f(x_t), g_t \rangle ] \nonumber \\
    =&  \lla \nabla f(x_t), \int_{\xi} \clip(\nabla f(x_t)+\xi,c) p(\xi) d \xi \rra \nonumber \\
    =&  \int_{\xi} \lla \nabla f(x_t),  \clip(\nabla f(x_t)+\xi,c) p(\xi) d \xi \rra \nonumber \\
     =&  \|\nabla f(x_t)\| \underbrace{\int_{\xi}  \|\clip(\nabla f(x_t)+\xi,c)\| cos(\nabla f(x_t), \nabla f(x_t) + \xi) p(\xi) d \xi}_{T_7} 
\end{align}
where the last equality is by definition of cosine and the fact that the clipping operation keeps directions.

In the following, we want to show $T_7$ is an non-decreasing function of  $\|\nabla f(x_t)\|$, then the result can be directly obtained from first part of the theorem.

Notice that $T_7$ is invariant to simultaneous rotation of $\nabla f(x_t)$ and the noise distribution (i.e., changing the coordinate axis of the space). Thus, wlog, we can assume $\nabla f(x_t)_1 = y > 0$ and $\nabla f(x_t)_i = 0$ for $2\leq i\leq d$. To show $T_7$ is a non-decreasing function of $\|\nabla f(x_t)\|$, it is enough to show each term in the integration is an non-decreasing function of $y$. {I.e., it is enough to show that, for all $\xi_t$, the following quantity} 
\begin{align}\label{eq: term_increase}
     \|\clip(\nabla f(x_t)+\xi_t,c)\| cos(\nabla f(x_t), \nabla f(x_t) + \xi_t)
\end{align}
is an non-decreasing function of $y$ for $y>0$ when $\nabla f(x_t)= [y,0,...,0]$ .

First consider the case where  $\|\nabla f(x_t)+\xi_t\| \leq c$. In this case, \eqref{eq: term_increase} reduces to
\begin{align}
     &\|\nabla f(x_t)+\xi_t\| cos(\nabla f(x_t), \nabla f(x_t) + \xi_t) \nonumber \\
     = & \|\nabla f(x_t)+\xi_t\| \frac{\langle \nabla f(x_t), \nabla f(x_t) + \xi_t \rangle }{\|\nabla f(x_t)\| \|\nabla f(x_t) + \xi_t\|} \nonumber \\
     = &  \frac{\langle \nabla f(x_t), \nabla f(x_t) + \xi_t \rangle }{\|\nabla f(x_t)\|} \nonumber \\
     = &  \frac{y(y+\xi_{t,1}) }{y}  
     =  y+\xi_{t,1}
\end{align}
which is a monotonically  increasing function of $y$. 

Now consider the case with $\|\nabla f(x_t)+\xi_t\| \geq c$, we have
\begin{align}\label{eq: clip_cos}
     & \|\clip(\nabla f(x_t)+\xi,c)\| cos(\nabla f(x_t), \nabla f(x_t) + \xi_t) \nonumber \\
     =& c \cdot cos(\nabla f(x_t), \nabla f(x_t) + \xi_t) \nonumber \\
     =& c \frac{\langle \nabla f(x_t), \nabla f(x_t) + \xi_t \rangle }{\|\nabla f(x_t)\| \|\nabla f(x_t) + \xi_t\|} \nonumber \\ 
     =& c \frac{y(y+\xi_{t,1}) }{y \sqrt{(y+\xi_{t,1})^2 + \sum_{i=2}^d \xi_{t,i}^2}} = c \frac{(y+\xi_{t,1}) }{ \sqrt{(y+\xi_{t,1})^2 + \sum_{i=2}^d \xi_{t,i}^2}} 
\end{align}
which is a non-decreasing function of $y$.

To see it is non-decreasing, define
\begin{align}
    r(z) = c \frac{z}{\sqrt{z^2+q^2}},
\end{align}
we have $ r'(z) =c( 1-\frac{z^2}{z^2+q^2}) \geq 0$.  The term in RHS of \eqref{eq: clip_cos} can be treated as $z = y+\xi_{t,1}$ and $q^2 = \sum_{i=2}^d \xi_{t,i}^2$.

Since the clipping function is continuous, combined with the above results, we know \eqref{eq: term_increase} is an non-decreasing function of $\|\nabla f(x_t)\|$.

Then we have
\begin{align}
    &\mathbb E[\langle \nabla f(x_t), g_t \rangle ] \nonumber \\
     =&  \|\nabla f(x_t)\| {\int_{\xi_t}  \|\clip(\nabla f(x_t)+\xi_t,c)\| cos(\nabla f(x_t), \nabla f(x_t) + \xi_t) p(\xi_t) d \xi_t}  \nonumber \\
     \geq &   \|\nabla f(x_t)\| {\int_{\xi_t}  \|\clip(\frac{3}{4} c\frac{\nabla f(x_t)}{\|\nabla f(x_t)\|}+\xi_t,c)\| cos(\frac{3}{4} c\frac{\nabla f(x_t)}{\|\nabla f(x_t)\|}, \frac{3}{4} c\frac{\nabla f(x_t)}{\|\nabla f(x_t)\|} + \xi_t) p(\xi_t) d \xi_t}  
\end{align}

From first part of the theorem, we know when $\|\nabla f(x_t)\| =\frac{3}{4} c$, we have \begin{align}
   \mathbb E[\langle \nabla f(x_t), g_t \rangle ] \geq  \| \nabla f(x_t)\|^2 \mathbb P(\|\xi_t\|<\frac{c}{4}) =  \| \nabla f(x_t)\| \left(\frac{3}{4}c \cdot \mathbb P(\|\xi_t\|<\frac{c}{4})\right)
\end{align}  
Combine the above result with \eqref{eq: T_7} and the non-decreasing property of $T_7$, we see that when $\|\nabla f(x_t)\| \geq \frac{3}{4}c$, the following holds:
$$ \| \nabla f(x_t)\|\left(\frac{3}{4}c \cdot \mathbb P(\|\xi\|<\frac{c}{4})\right)\le \mathbb E[\langle \nabla f(x_t),g_t \rangle ]=  \| \nabla f(x_t)\| T_7,$$
which implies 
$T_7\ge \frac{3}{4}c \cdot \mathbb P(\|\xi\|<\frac{c}{4}).$ Substituting this lower bound into \eqref{eq: T_7} finishes the proof. \qed

\subsection{Technical lemmas}
\begin{lemma} \label{lem: cos}
For any $g$ and $\xi$, we have 
\begin{align}
    cos(g, g + \xi)+  cos(g, g- \xi) \geq 0
\end{align}
\end{lemma}
\textbf{Proof:}
By definition of $cos$, we have
\begin{align} \label{eq: cos_reduce}
    &cos(g, g + \xi)+  cos(g, g- \xi) \nonumber \\
    =&\frac{\langle g,g+\xi \rangle}{\|g\| \|g+\xi\|} + \frac{\langle g, g- \xi \rangle}{\|g\|\|g-\xi\|} \nonumber \\
    =& \frac{
\|g\|}{\|g+\xi\|}  + \frac{
\|g\|}{\|g-\xi\|}  +    \frac{\langle g,\xi \rangle}{\|g\| \|g+\xi\|} - \frac{\langle g,  \xi \rangle}{\|g-\xi\|} \nonumber \\
=& \frac{
\|g\|}{\|g+\xi\|}  + \frac{
\|g\|}{\|g-\xi\|}  +    \frac{\|\xi\| cos(g,\xi) }{\|g+\xi\|} - \frac{\|\xi\|cos(g,\xi)}{\|g-\xi\|} \nonumber \\
=& \frac{\|g+\xi\|(\|g\|- \|\xi\|e) +\|g-\xi\|(\|g\|+ \|\xi\|e) }{\|g+\xi\|\|g-\xi\|}
\end{align}
where $e = cos(g,\xi)$.

To prove the desired result, we only need the numerator of RHS of 
\eqref{eq: cos_reduce} to be non-negative.

Denote $h(\xi) = cos(g, g + \xi)+  cos(g, g- \xi)$, since $h$ is rotation invariant, we can assume $\xi_1 >0$ and $\xi_{t,i} =0$ for $2\leq i \leq d$ wlog. Also, because $h(\xi) = h(-\xi)$, we can assume $g_1 \geq 0$ wlog.

Now suppose $g_1 = a$, $\sum_{i=2}^d g_i^2 = b^2$, Denote the numerator of RHS of 
\eqref{eq: cos_reduce} as $T_3$, it can be written as 
\begin{align}
    T_3 =& \|g+\xi\|(\|g\|- \|\xi\|e) +\|g-\xi\|(\|g\|+ \|\xi\|e)  \nonumber \\
    = &\underbrace{\sqrt{b^2 + (a+\xi_1)^2}(\sqrt{a^2+b^2}-\xi_1e)}_{T_4} + \underbrace{ \sqrt{b^2 + (a-\xi_1)^2}(\sqrt{a^2+b^2}+\xi_1e) }_{T_5}
\end{align}
and $e = \frac{\langle g,\xi\rangle}{\|g\| \|\xi\|} =  \frac{a}{\sqrt{a^2+b^2}}$.

Now let us analyze when $T_3$ can be possibly less than 0.  Recall that by assumption, $\xi_1>0$ and $e \geq 0$. Then we know  $T_5 \geq 0$. We have $T_3 \geq 0$ trivially when $T_4 \geq 0$, i.e. when $\xi_1e \leq \sqrt{a^2 + b^2}$. 

Now assume $\xi_1e > \sqrt{a^2 + b^2}$. To ensure $T_3 \geq 0$, we can alternatively ensure $T_5^2 \geq T_4^2$ in this case.

We have 
\begin{align}
    T_5^2 - T_4^2 = & {({b^2 + (a-\xi_1)^2})(\sqrt{a^2+b^2}+\xi_1e)^2} - { ({b^2 + (a+\xi_1)^2})(\sqrt{a^2+b^2}-\xi_1e)^2 } \nonumber \\
    = & 4b^2\xi_1e\sqrt{a^2+b^2} + \underbrace{4\xi_1e \sqrt{a^2+b^2}(a^2+\xi_1^2) - 4a\xi_1(a^2+b^2+\xi_1^2e^2)}_{T_6}
\end{align}

For $T_6$, we can further simplify it as
\begin{align}
    T_6 = & 4\xi_1e \sqrt{a^2+b^2}(a^2+\xi_1^2) - 4a\xi_1(a^2+b^2+\xi_1^2e^2) \nonumber \\
    = &4\xi_1a(a^2+\xi_1^2) - 4a\xi_1(a^2+b^2+\xi_1^2e^2) \nonumber \\
    = &  4\xi_1a (\xi_1^2(1-e^2) - b^2) \nonumber \\
    = &  4\xi_1a (\xi_1^2(\frac{b^2}{a^2+ b^2}) - b^2) \nonumber \\
    = &  4\xi_1a (b^2(\frac{\xi_1^2 - (a^2+b^2)}{a^2+ b^2}) ) \nonumber \\
    \geq & 0
\end{align}
where the last inequality is because $\xi^2 \geq \xi^2e^2 \geq a^2+b^2$ and $\xi_1a >0$ as assumed previously.

Combining all above, we have $T_6 \geq 0 \implies T_5^2- T_4^2 \geq 0 \implies T_3 \geq 0 \implies cos(g, g + \xi)+  cos(g, g- \xi) \geq 0$
 which proves the desired result. \qed

\begin{lemma}\label{lem: norm_to_cos}
For any $g$ and $\xi$, we have 
\begin{align}
    \|g + \xi\| \geq \|g - \xi\| 
\end{align}
if $cos(g, \xi) > 0$ and 
\begin{align}
    \|g + \xi\| \leq \|g - \xi\| 
\end{align}
if $cos(g, \xi) < 0$ .
\end{lemma}
\textbf{Proof:}
Express $\xi$ using a coordinate system with one axis parallel to $g$.  Define the basis of this coordinate system as $v_1, v_2,... v_d$ with $v_1$ = $g/\|g\|$. Then we have $\xi = \sum_{i=1}^d b_i v_i $ and $cos(g,\xi) > 0 $ if and only if $b_1 >0$.

In addition, we have
\begin{align}
    \|g+\xi\| =\sqrt{(\|g\| +b_1)^2 + \sum_{i=2}^d b_i^2 }
\end{align}
and 
\begin{align}
    \|g-\xi\| =\sqrt{(\|g\| -b_1)^2 + \sum_{i=2}^d b_i^2 }.
\end{align}

Then it is clear that $\|g+\xi\| \geq \|g-\xi\|$ when $b_1 >0$ which means $cos(g,\xi) > 0 $.

Similar arguments applies to the case with $cos(g,\xi) < 0 $ \qed
\section{Proof of Theorem \ref{thm: multimodal}}
\begin{reptheorem}{thm: multimodal}
Given $m$ distributions with the pdf of the $i$th distribution being  $p_i(\xi_t) = \phi_i(\|\xi_t - u_i\|)$ for some function $\phi_i$. If $\nabla f(x_t) = \sum_{i=1}^m w_iu_i$ for some $w_i \geq 0,  \sum_{i=1}^m w_i = 1$. Let $p'(\xi_t) = \sum_{i=1}^m w_i p_i(\xi_t - \nabla f(x_t) ), $ be a mixture of these distributions with  zero mean. If $\langle u_i, \nabla f(x_t) \rangle \geq 0, \forall i \in [m]$, we have 
\begin{align*}
\mathbb E_{\xi_t \sim {p'}}[\langle \nabla f(x_t), g_t \rangle ]  \geq \| \nabla f(x_t)\| \sum_{i=1}^m w_i\min(\|u_i\|,\frac{3}{4}c)\cos (\nabla f(x_t),u_i)  \mathbb P_{\xi_t \sim p_i}(\|\xi_t\|<\frac{c}{4}) \geq 0
\end{align*}
\end{reptheorem}

\textbf{Proof:}
First, we notice that Theorem \ref{thm: symmetric_descent} can be restated into a more general form as follows.

\begin{reptheorem}{thm: symmetric_descent}[restated]
Given a random variable $\xi \sim \tilde p$ with $\tilde p(\xi) = \tilde p(-\xi)$ being a symmetric distribution, for any vector $g$, we have
\begin{align}
\textrm{1. If } \|g\| \leq \frac{3}{4}c,&\quad \text{ then }\quad
\mathbb E[\langle g, \clip(g+\xi,c) \rangle ]  \geq \| g\|^2  \mathbb P\left(\|\xi\|<\frac{c}{4}\right)\\
\textrm{2. If } \|g\| > \frac{3}{4}c,&\quad \text{ then } \quad \mathbb E[\langle g,\clip(g+\xi,c)  \rangle ]  \geq \frac{3}{4}  c \| g\|  \mathbb P\left(\|\xi\|<\frac{c}{4} \right)
\end{align}
\end{reptheorem}
In addition, by sphere symmetricity, if $\xi \sim \hat p$ with $\hat p$  being a spherical distribution $\hat p(\xi) = \phi (\|\xi\|)$ for some function $\phi$, for any vector $g$, we have $\mathbb E [\clip (g + \xi)] = r g $ with $r$ being a constant (i.e. the expected clipped gradient is in the same direction as $g$). Combining with restated Theorem \ref{thm: symmetric_descent} above, we have when $\tilde p$ is a spherical distribution  with $\tilde p(\xi) = \phi (\|\xi\|)$, 
\begin{align}\label{eq: same_direc}
    \mathbb E[\clip(g+\xi,c)] = r g  
\end{align}
with $r \geq 0$ and
\begin{align}
    r\|g\| \geq \min(\frac{3}{4}c,\|g\|)\mathbb P\left(\|\xi\|<\frac{c}{4} \right).
\end{align}

Now we can use the above results to prove the theorem.

The expectation can be splitted as 
\begin{align}
    \mathbb E_{\xi_t \sim {p'}}[\langle \nabla f(x_t), g_t \rangle ] =  \sum_{i=1}^m w_i \mathbb E_{\xi_t \sim {p_i}}[\langle \nabla f(x_t), g_t \rangle ].
\end{align}
Then, because \eqref{eq: same_direc} and $ E_{\xi_t \sim {p_i}}[g_t] = u_i$ and that $p_i$ corresponds to a noise with spherical distribution added to $u_i$, we have
\begin{align}
    E_{\xi_t \sim {p_i}}[\langle \nabla f(x_t), g_t \rangle ] = \langle \nabla f(x_t),   E_{\xi_t \sim {p_i}}[g_t] \rangle  = \langle \nabla f(x_t),  r_i u_i \rangle 
\end{align}
with $r_i \|u_i\| \geq \min(\frac{3}{4}c,\|u_i\|)\mathbb P_{\xi_t \sim p_i}\left(\|\xi_t\|<\frac{c}{4} \right) $. Since we assumed $ \langle u_i, \nabla f(x_t) \rangle \geq 0$, we have 
\begin{align}
    \mathbb E_{\xi_t \sim {p'}}[\langle \nabla f(x_t), g_t \rangle ] \geq   \|\nabla f(x_t)\|  \sum_{i=1}^m w_i\min(\frac{3}{4}c,\|u_i\|) \cos (u_i,\nabla f(x_t))\mathbb P_{\xi_t \sim p_i}\left(\|\xi_t\|<\frac{c}{4} \right) \geq 0
\end{align}
which is the desired result. \qed

\section{Proof of Theorem \ref{thm: dp_conv}} 
Recall the algorithm has the following update rule
\begin{align}
x_{t+1} = x_t - \alpha \left( \left( \frac{1}{|S_t|} \sum_{i\in S_t} \clip(\nabla f(x_t) + \xi_{t,i},c) \right) + Z_t\right)
\end{align}
where $g_{t,i} \triangleq \nabla f(x_t)+ \xi_{t,i}$ is the stochastic gradient at iteration $t$ evaluated on sample $i$, and $ S_t$ is a subset of whole dataset $D$; 
$Z_t \sim \mathcal N(0, \sigma^2  I)$ is the noise added for privacy. We denote $g_t = \frac{1}{|S_t|}\sum_{i\in  S_t} \clip(\nabla f(x_t)+ \xi_{t,i},c)$ in the remaining parts of the  proof to simplify notation.

Following traditional convergence analysis of SGD using smoothness assumption, we first have
\begin{align}
    f(x_{t+1}) \leq f(x_t) + \langle \nabla f(x_t), x_{t+1} - x_t \rangle + \frac{1}{2}G \|x_{t+1} - x_t\|^2
\end{align}
which translates into
\begin{align}
    f(x_{t+1}) \leq f(x_t) -  \alpha \langle \nabla f(x_t), (  g_t  + Z_t) \rangle + \frac{1}{2}G \alpha^2 \| g_t + Z_t\|^2
\end{align}

Taking expectation conditioned on $x_t$, we have
\begin{align}
    &\mathbb E[f(x_{t+1})]  \nonumber \\
    \leq & f(x_t) -  \alpha \mathbb E[ \langle \nabla f(x_t),   g_t   \rangle]+ \frac{1}{2}G \alpha^2 (\mathbb E[\| g_t\|^2]  + \sigma^2 c^2 d) \nonumber \\
    \leq & f(x_t) -  \alpha  \mathbb E[  \langle \nabla f(x_t),   g_t   \rangle]+ \frac{1}{2}G \alpha^2 (c^2  + \sigma^2 c^2 d).
\end{align}

Take overall expectation and sum over $t \in [T]$ and rearrange, we have 
\begin{align}
    \sum_{t=1}^T \alpha  \mathbb E[ \langle \nabla f(x_t),   g_t  \rangle]
    \leq & f(x_1) - \mathbb E[f(x_{T+1})]  + T \frac{1}{2}G \alpha^2 (c^2  + \sigma^2 d).
\end{align}

Dividing both sides by $T\alpha$, we get
\begin{align}
    \frac{1}{T}\sum_{t=1}^T  \mathbb E[ \langle \nabla f(x_t),  g_t \rangle]
    \leq & \frac{f(x_1) - \mathbb E[f(x_{T+1})]]}{T\alpha}  +  \frac{1}{2}G \alpha (c^2  + \sigma^2 d).
\end{align}

To achieve $(\epsilon,\delta)$-privacy, we need $\sigma^2 = v\frac{Tc^2\ln(\frac{1}{\delta})}{n^2 \epsilon^2}$ for some constant $v$ by Theorem 1 in \cite{abadi2016deep}. Substituting the expression of $\sigma^2$ into the above inequality, we get
\begin{align}
    \frac{1}{T}\sum_{t=1}^T \langle  \mathbb E[ \nabla f(x_t),   g_t   \rangle]
    \leq & \frac{D_f}{T\alpha }  +  \frac{1}{2}G \alpha (c^2  + v\frac{T\ln(\frac{1}{\delta})}{n^2 \epsilon^2}c^2  d)
\end{align}
where we define $D_f = f(x_1) - \min_x f(x)$.

Setting $T\alpha = \frac{\sqrt{D_f}n\epsilon}{
\sqrt{G}c\sqrt{d}\sqrt{\ln(\frac{1}{\delta})}}$, we have
\begin{align}
    \frac{1}{T}  \sum_{t=1}^T \mathbb E[ \langle \nabla f(x_t),   g_t   \rangle]
    \leq & 
    \left(\frac{1}{2 }v+1\right) \frac{c \sqrt{D_fGd\ln(\frac{1}{\delta})}
}{n\epsilon }  +  \frac{1}{2}G \alpha c^2.
\end{align}

Setting $\alpha = \frac{ \sqrt{D_fd\ln(\frac{1}{\delta})}
}{n\epsilon c\sqrt{G} }$, we have
\begin{align}\label{eq: sgd_clip}
    \frac{1}{T}\sum_{t=1}^T   \mathbb E[ \langle \nabla f(x_t),  g_t   \rangle]
    \leq & 
    \left(\frac{1}{2 }v+\frac{3}{2}\right)\frac{c \sqrt{D_fGd\ln(\frac{1}{\delta})}
}{n\epsilon }.
\end{align}

The remaining step is to analyze the term on LHS of \eqref{eq: sgd_clip}. We first notice that the gradient sampling scheme yields
\begin{align}
     \mathbb E[ \langle \nabla f(x_t),   g_t \rangle] =  \mathbb E_{\xi_t \sim p}[ \langle \nabla f(x_t),   \clip(\nabla f(x_t)+ \xi_t,c)   \rangle]
\end{align} 
with $\xi_t$ being a discrete random variable that can takes values $\xi_{t,i}, i \in D $ with equal probability and $D$ is the whole dataset. 

Now it is time to split the bias as following. 
\begin{align*}
    \mathbb E_{\xi_t \sim p}[\langle \nabla f(x_t), g_t \rangle ] 
    =& \mathbb E_{\xi_t \sim \tilde{p}} [\langle \nabla f(x_t),g_t \rangle] +\int \langle \nabla f(x_t),\clip(\nabla f(x_t)+ \xi_t,c) \rangle (p_t(\xi_t)-\tilde{p_t}( \xi_t)) d\xi_t 
\end{align*}
with $\tilde p$ being a symmetric distribution. Applying Theorem \ref{thm: symmetric_descent}, we have
\begin{align}
    \mathbb E_{\xi_t \sim \tilde p}[ \langle \nabla f(x_t),   g_t,   \rangle] \geq \mathbb P_{\xi_t \sim \tilde p}(\|\xi_t\|<\frac{c}{4}) \| \nabla f(x_t)\|^2  
\end{align}
when $\|\nabla f(x_t)\| \leq \frac{3}{4}c$ and
\begin{align}
    \mathbb E_{\xi_{t} \sim \tilde p}[ \langle \nabla f(x_t),   g_t  \rangle] \geq \frac{3}{4} \mathbb P_{\xi_{t} \sim \tilde p}(\|\xi_t\|<\frac{c}{4}) c \| \nabla f(x_t)\| 
\end{align}
when $\|\nabla f(x_t)\| \geq \frac{3}{4}c$.

Now we bound the bias term using Wasserstein distance as follows. 
{\small
\begin{align}
&-\int \langle \nabla f(x_t),\clip(\nabla f(x_t)+ \xi_t,c) \rangle (p_t(\xi_t)-\tilde{p_t}( \xi_t)) d\xi_t \nonumber \\
   = &
\int \langle \nabla f(x_t),\clip(\nabla f(x_t)+ \xi_t,c) \rangle (\tilde p(\xi_t)-{p}(\xi_t)) d\xi_t \nonumber \\
=& \int \langle \nabla f(x_t),\clip(\nabla f(x_t)+ \xi_t,c) \rangle \tilde p(\xi_t) d\xi_t - \int \langle \nabla f(x_t),\clip(\nabla f(x_t)+ \xi_t',c) \rangle  p(\xi_t') d\xi_t'  \nonumber \\
=& \int \int (\langle \nabla f(x_t),\clip(\nabla f(x_t)+ \xi_t,c) \rangle - \langle \nabla f(x_t), \clip(\nabla f(x_t)+ \xi_t',c) \rangle) \gamma (\xi_t,\xi_t')  d\xi_t d\xi'_t \nonumber \\
\leq & \int \int |\langle \nabla f(x_t),\clip(\nabla f(x_t)+ \xi_t,c) \rangle - \langle \nabla f(x_t), \clip(\nabla f(x_t)+ \xi_t',c) \rangle| \gamma (\xi_t,\xi_t')  d\xi_t d\xi'_t
\end{align}
}%
where $\gamma$ is any joint distribution with marginal $\tilde p$ and $ p$. Thus, we have 
\begin{align*}
&-\int \langle \nabla f(x_t),\clip(\nabla f(x_t)+ \xi_t,c) \rangle (p_t(\xi_t)-\tilde{p_t}( \xi_t)) d\xi_t \nonumber \\
 \leq & \inf_{\gamma \in \Gamma(\tilde p, p)}\int \int |\langle \nabla f(x_t),\clip(\nabla f(x_t)+ \xi_t,c) \rangle - \langle \nabla f(x_t), \clip(\nabla f(x_t)+ \xi_t',c) \rangle| \gamma (\xi_t,\xi_t')  d\xi_t d\xi'_t 
\end{align*}
where $\Gamma(\tilde p,  p)$ is the set of all coupling with marginals $\tilde p$ and $ p$ on the two factors, respectively. If we define the distance function $d_{y,c}(a,b) = |\langle y, \clip(y+ a,c) \rangle - \langle y, \clip(y+ b,c) \rangle|$, we have 
{\small
\begin{align}
   & -\int \langle \nabla f(x_t),\clip(\nabla f(x_t)+ \xi_t,c) \rangle (p_t(\xi_t)-\tilde{p_t}( \xi_t)) d\xi_t \nonumber \\
    \leq &\inf_{\gamma \in \Gamma(\tilde p, p)}\int \int d_{\nabla f(x_t), c}(\xi_t, \xi_t') \gamma (\xi_t,\xi_t')  d\xi_t d\xi'_t  = W_{\nabla f(x_t),c} (\tilde p_t,p_t)
\end{align}
}%
which we define $W_{v,c}( p, p')$ as the Wasserstein distance between $ p$ and $p'$ using the metric $d_{v,c}$.

Wrapping up, define 
\[
    h(y) = \bigg\{\begin{array}{lr}
        y^2, & \text{for } y \leq 3c/4\\
        \frac{3}{4} c y, & \text{for } y > 3c/4
        \end{array},
  \]
  we have
\begin{align}\label{eq: sgd_clip_final}
    \frac{1}{T}\sum_{t=1}^T \mathbb P_{\xi_{t} \sim \tilde p_t}(\|\xi_t\|<\frac{c}{4}) h(\|\nabla f(x_t)\|)
    \leq & 
    \left(\frac{1}{2 }v+\frac{3}{2}\right)\frac{c \sqrt{D_fGd\ln(\frac{1}{\delta})} 
}{n\epsilon } + \frac{1}{T} \sum_{t=1}^T W_{\nabla f(x_t),c} (\tilde p_t,p_t)
\end{align}
which is the desired result. \hfill $\square$

\section{Proof of Theorem \ref{thm: scale_noise_main}}
\begin{reptheorem}{thm: scale_noise_main}
Let $g_t = \clip(\nabla f(x_t)+ \xi_t + k \zeta_t,c)$ and $\zeta_t \sim \mathcal N(0,  I)$. Then gradient clipping algorithm has following properties: 
\begin{align}
\mathbb E_{\xi_t \sim {p}, \zeta_t}[\langle \nabla f(x_t), g_t \rangle ]  \geq \| \nabla f(x_t)\| \min\left\{\| \nabla f(x_t)\|, \frac{3}{4}c\right\}  \mathbb P(\|k\zeta_t\|<\frac{c}{4}) - O(\frac{\sigma^2_{\xi_t}}{k^2})
\end{align} 
where $\sigma^2_{\xi_t}$ is the variance of the gradient noise $\xi_t$.
\end{reptheorem}
\textbf{Proof}: Define $W_t =  \xi_t + k \zeta_t $ be the total noise on the gradient before clipping and $W_t \sim  \bar p $. We know $\mathbb E [W_t] = 0$ and $\bar p(W_t) = \int_{\xi_t} p(\xi_t) \frac{1}{k}\psi(\frac{W_t - \xi_t}{k}) d\xi_t$ with $\psi$ being the pdf of $\mathcal N(0,  I)$. The proof idea is to bound the total variation distance between $\bar p(W_t)$ and $\frac{1}{k}\psi$ as $O(\frac{\sigma_{\xi_t}^2}{k^2})$, then use this distance to bound the clipping bias $b_t$. This implies $\bar p(W_t)$ will become more and more symmetric as $k$ increases.

We have 
\begin{align}
   &\int \left|\bar p(W_t) - \frac{1}{k}\psi(\frac{W_t}{k})\right| d{W_t} \nonumber \\ 
   = & \int_{W_t} \left|\int_{\xi_t} p(\xi_t) \frac{1}{k}\psi(\frac{W_t - \xi_t}{k}) d\xi_t- \frac{1}{k}\psi(\frac{W_t}{k})\right| d{W_t} \nonumber \\
    = & k\int_{W'_t} \left|\int_{\xi_t} p(\xi_t) \frac{1}{k}\psi(W_t'-\frac{ \xi_t}{k}) d\xi_t- \frac{1}{k}\psi(W'_t)\right| d{W'_t} 
\end{align}

By Taylor's series, we have
\begin{align}\label{eq:taylor}
    \psi(W_t'-\frac{ \xi_t}{k}) =   \psi({W'_t}) +   \langle \nabla \psi({W'_t}), \frac{- \xi_t}{k}\rangle +  \int_{0}^1 \left\langle \frac{\xi_t}{k}, \nabla^2 \psi(W_t'-\tau \frac{\xi_t}{k}) \frac{\xi_t}{k} \right\rangle(1-\tau) d \tau
\end{align}

Then, 
\begin{align}\label{eq: taylor_remainder}
   &\int |\bar p(W_t) - \frac{1}{k}\psi(\frac{W_t}{k})| d{W_t} \nonumber \\ 
    = & \int_{W'_t} \left|\int_{\xi_t} p(\xi_t) \psi(W_t'-\frac{ \xi_t}{k}) d\xi_t- \psi(W'_t)\right| d{W'_t} \nonumber \\
    = & \int_{W'_t} \left|\int_{\xi_t} p(\xi_t)  \int_{0}^1 \langle \frac{\xi_t}{k}, \nabla^2 \psi(W_t'-t \frac{\xi_t}{k}) \frac{\xi_t}{k} \rangle(1-\tau) d t d \xi_t\right| d{W'_t} \nonumber \\
    \leq & \int_{0}^1  \int_{\xi_t} \int_{W'_t} \left| p(\xi_t)  \langle \frac{\xi_t}{k}, \nabla^2 \psi(W_t'-t \frac{\xi_t}{k}) \frac{\xi_t}{k} \rangle(1-\tau) \right| d{W'_t} d \xi_t d \tau
\end{align} 
{where the second equality is obtained by applying \eqref{eq:taylor} and using the fact that $\xi_t$ is zero mean.}

{Noticing that $\tau \leq 1$ and define $\hat W_t = W_t'-\tau \frac{\xi_t}{k}$, we have
\begin{align}
  &  \int_{W'_t} \left| p(\xi_t)  \langle \frac{\xi_t}{k}, \nabla^2 
\psi(W_t'-\tau \frac{\xi_t}{k}) \frac{\xi_t}{k} \rangle(1-\tau) \right| d{W'_t} \nonumber \\
= &   p(\xi_t) (1-\tau)    \int_{ \hat W_t} \left|\langle \frac{\xi_t}{k}, \nabla^2 
\psi(\hat W_t') \frac{\xi_t}{k} \rangle \right| \frac{d{W'_t}}{d{\hat W_t}} d{\hat W_t}  \nonumber \\
= &   p(\xi_t) (1-\tau)    \int_{ \hat W_t} \left|\langle \frac{\xi_t}{k}, \nabla^2 
\psi(\hat W_t) \frac{\xi_t}{k} \rangle \right| d{\hat W_t} 
\end{align}
and the integration term only depends on $\|\frac{\xi_t}{k}\|$ due to sphere symmetricity of $\psi$.} Thus we can assume $\xi_{t,1} = \|\xi_t\|$ and $\xi_{t,i} = 0$ for $i \geq 2$, wlog. Then, we have
\begin{align}\label{eq: int_1}
    &\int_{W'_t} \left| p(\xi_t)  \langle \frac{\xi_t}{k}, \nabla^2 \psi(W_t'-\tau \frac{\xi_t}{k}) \frac{\xi_t}{k} \rangle(1-\tau) \right| d{W'_t} \nonumber \\
    \leq  & p(\xi_t) (1-\tau) \int_{W'_t} \frac{\|\xi_t\|^2}{k^2} \left|     \nabla^2_{1,1} \psi(W_t'-\tau\frac{\xi_t}{k})   \right| d{W'_t} \nonumber \\
     \leq  & p(\xi_t) (1-\tau) \int_{\hat W_t} \frac{\|\xi_t\|^2}{k^2} \left|     \nabla^2_{1,1} \psi(W_t)   \right| \frac{d{W'_t}}{d{\hat W_t}} d{\hat W_t} \nonumber \\
    \leq  & p(\xi_t) (1-\tau)  \frac{\|\xi_t\|^2}{k^2} q
\end{align}
where we have define $\hat W_t= W_t'-\tau \frac{\xi_t}{k}$ and  $q = \int_{-\infty}^{\infty} |h''(x)| d x $ with $h(x)$ being the pdf of 1-dimensional standard normal distribution. Thus, $q$ is a dimension independent constant.

Substituting \eqref{eq: int_1} into \eqref{eq: taylor_remainder}, we get
\begin{align}\label{eq: total_var}
   &\int |\bar p(W_t) - \frac{1}{k}\psi(\frac{W_t}{k})| d_{W_t} \nonumber \\ 
    \leq & \int_{0}^1  \int_{\xi_t} \int_{W'_t} \left| p(\xi_t)  \langle \frac{\xi_t}{k}, \nabla^2 \psi(W_t'-\tau \frac{\xi_t}{k}) \frac{\xi_t}{k} \rangle(1-\tau) \right| d{W'_t} d \xi_t d \tau \nonumber \\
    \leq &\int_{0}^1  \int_{\xi_t} p(\xi_t) (1-\tau)  \frac{\|\xi_t\|^2}{k^2} q d \xi_t d \tau \nonumber \\
    = &\int_{0}^1 (1-\tau)  \frac{\sigma_{\xi_t}^2}{k^2} q  d \tau \nonumber \\
    = &\frac{1}{2} \frac{\sigma_{\xi_t}^2}{k^2} q   
\end{align} 
where we used the fact that $\mathbb E [\xi_t] = 0$ and defined $\sigma_{\xi_t}^2$ being the variance of $\xi_t$.

By \eqref{eq: split_bias}, we know

\begin{align}\label{eq: bias_split_2}
    \mathbb E_{\xi_t \sim p, \zeta_t}[\langle \nabla f(x_t), g_t \rangle ]
     & = \mathbb E_{W_t \sim \tilde{p}} [\langle \nabla f(x_t),g_t \rangle] \nonumber\\
    & + \underbrace{\int \langle \nabla f(x_t),\clip(\nabla f(x_t)+ W_t,c) \rangle (p_t(W_t)-\tilde{p_t}( W_t)) d W_t }_{b_t}
\end{align}

Let $\tilde p $ be the pdf of $k\zeta_t$, from Theorem \ref{thm: symmetric_descent}, we have
\begin{align}\label{eq: descent_correct}
    \mathbb E_{W_t \sim \tilde{p}} [\langle \nabla f(x_t),g_t \rangle] \geq \|\nabla f(x_t)\| \min\left\{\frac{3}{4}c, \|\nabla f(x_t)\| \right\}\mathbb P (\|k\zeta_t\| \leq \frac{c}{4})
\end{align}

In addition, we can bound $b_t$ as 
\begin{align}\label{eq: bias_reduce}
    |b_t| \leq   \|\nabla f(x_t)\| c  \int |p_t(\xi_t)-\tilde{p_t}( \xi_t)| d\xi_t \leq \|\nabla f(x_t)\|  \frac{c}{2} \frac{\sigma_{\xi_t}^2}{k^2} q = O(\frac{\sigma_{\xi_t}^2}{k^2})
\end{align}
by \eqref{eq: total_var}.

Combining \eqref{eq: bias_split_2}, \eqref{eq: bias_reduce},  and \eqref{eq: descent_correct} finishes the proof. \qed

\section{More experiments on random projection}
\begin{figure}[htbp]
	\centering
	\begin{minipage}[c]{0.24\linewidth}
		\includegraphics[width=\linewidth]{figures/mnist/epoch0/mnist_gradient_0_0.png}
		\centering{\small (a) Repeat 1}
	\end{minipage}
	\begin{minipage}[c]{0.24\linewidth}
		\includegraphics[width=\linewidth]{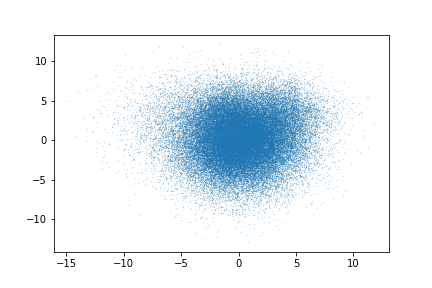}
		\centering{\small (b) Repeat 2}
	\end{minipage}
	\begin{minipage}[c]{0.24\linewidth}
		\includegraphics[width=\linewidth]{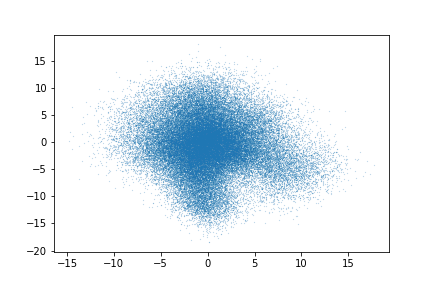}
		\centering{\small (c) Repeat 3}
	\end{minipage}
	\begin{minipage}[c]{0.24\linewidth}
		\includegraphics[width=\linewidth]{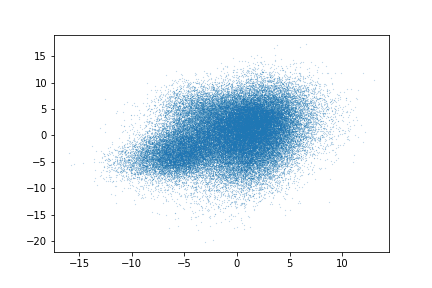}
		\centering{\small (d) Repeat 4}
	\end{minipage}
	\\ 
		\begin{minipage}[c]{0.24\linewidth}
		\includegraphics[width=\linewidth]{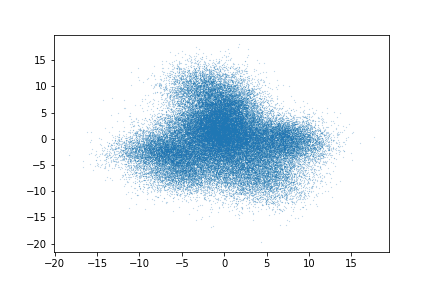}
		\centering{\small (e) Repeat 5}
	\end{minipage}
	\begin{minipage}[c]{0.24\linewidth}
		\includegraphics[width=\linewidth]{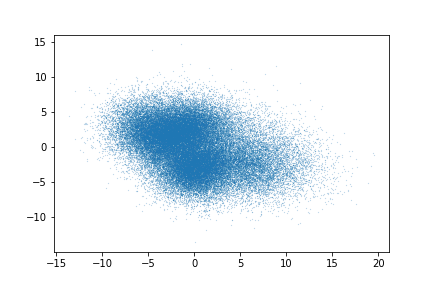}
		\centering{\small (f) Repeat 6}
	\end{minipage}
	\begin{minipage}[c]{0.24\linewidth}
		\includegraphics[width=\linewidth]{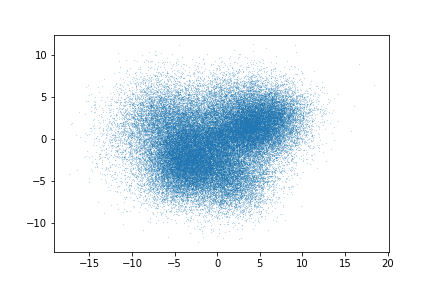}
		\centering{\small (g) Repeat 7}
	\end{minipage}
	\begin{minipage}[c]{0.24\linewidth}
		\includegraphics[width=\linewidth]{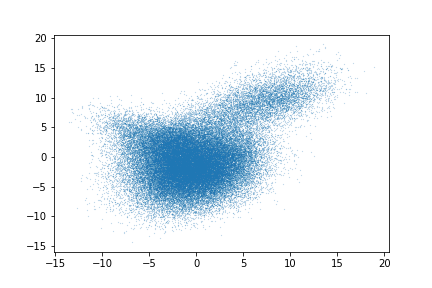}
		\centering{\small (h) Repeat 8}
	\end{minipage}
	\caption{\small Distribution of gradients on MNIST after  epochs 0 projected using different random matrices. 
	}
    \label{fig:mnist_ep0}
\end{figure}

\begin{figure}[htbp]
	\centering
	\begin{minipage}[c]{0.24\linewidth}
		\includegraphics[width=\linewidth]{figures/mnist/epoch3/mnist_gradient_3_0.png}
		\centering{\small (a) Repeat 1}
	\end{minipage}
	\begin{minipage}[c]{0.24\linewidth}
		\includegraphics[width=\linewidth]{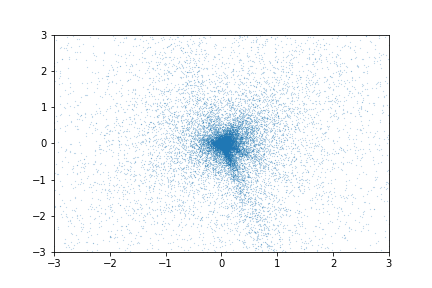}
		\centering{\small (b) Repeat 2}
	\end{minipage}
	\begin{minipage}[c]{0.24\linewidth}
		\includegraphics[width=\linewidth]{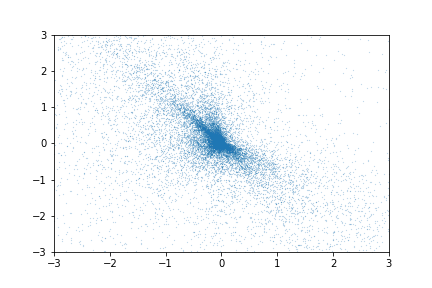}
		\centering{\small (c) Repeat 3}
	\end{minipage}
	\begin{minipage}[c]{0.24\linewidth}
		\includegraphics[width=\linewidth]{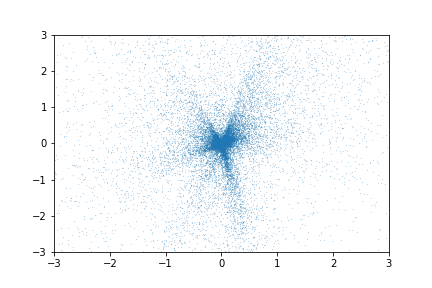}
		\centering{\small (d) Repeat 4}
	\end{minipage}
	\\ 
		\begin{minipage}[c]{0.24\linewidth}
		\includegraphics[width=\linewidth]{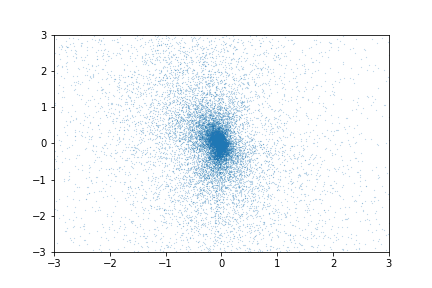}
		\centering{\small (e) Repeat 5}
	\end{minipage}
	\begin{minipage}[c]{0.24\linewidth}
		\includegraphics[width=\linewidth]{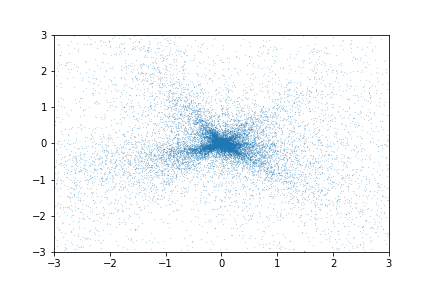}
		\centering{\small (f) Repeat 6}
	\end{minipage}
	\begin{minipage}[c]{0.24\linewidth}
		\includegraphics[width=\linewidth]{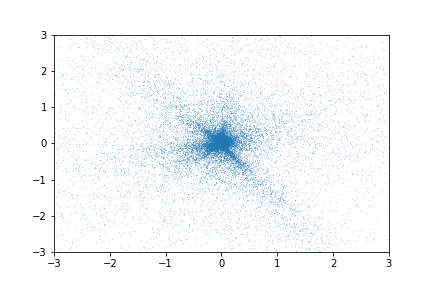}
		\centering{\small (g) Repeat 7}
	\end{minipage}
	\begin{minipage}[c]{0.24\linewidth}
		\includegraphics[width=\linewidth]{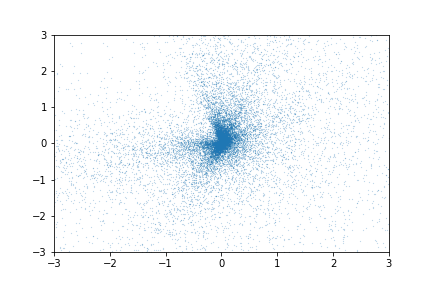}
		\centering{\small (h) Repeat 8}
	\end{minipage}
	\caption{\small Distribution of gradients on MNIST after  epochs 3 projected using different random matrices. 
	}
    \label{fig:mnist_ep3}
\end{figure}

\begin{figure}[htbp]
	\centering
	\begin{minipage}[c]{0.24\linewidth}
		\includegraphics[width=\linewidth]{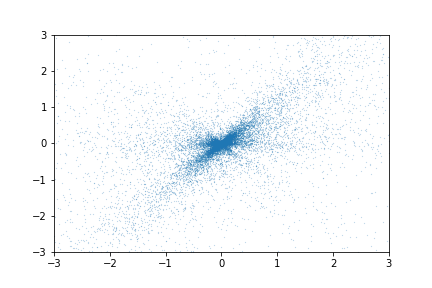}
		\centering{\small (a) Repeat 1}
	\end{minipage}
	\begin{minipage}[c]{0.24\linewidth}
		\includegraphics[width=\linewidth]{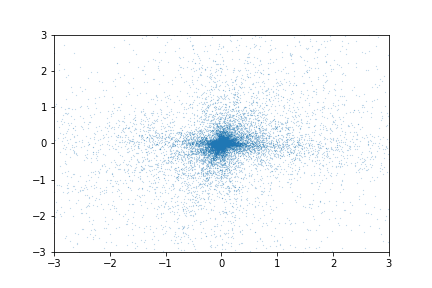}
		\centering{\small (b) Repeat 2}
	\end{minipage}
	\begin{minipage}[c]{0.24\linewidth}
		\includegraphics[width=\linewidth]{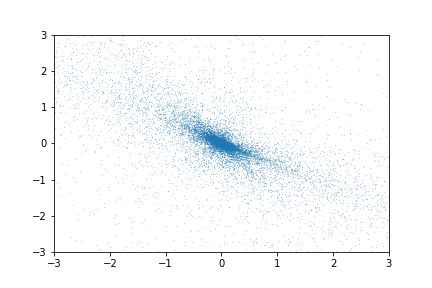}
		\centering{\small (c) Repeat 3}
	\end{minipage}
	\begin{minipage}[c]{0.24\linewidth}
		\includegraphics[width=\linewidth]{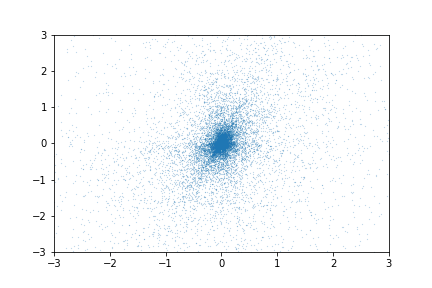}
		\centering{\small (d) Repeat 4}
	\end{minipage}
	\\ 
		\begin{minipage}[c]{0.24\linewidth}
		\includegraphics[width=\linewidth]{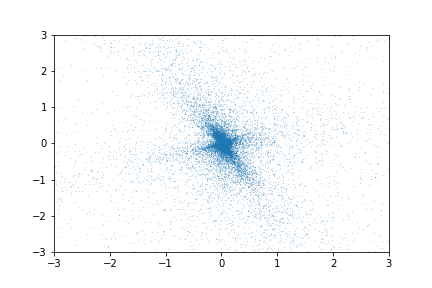}
		\centering{\small (e) Repeat 5}
	\end{minipage}
	\begin{minipage}[c]{0.24\linewidth}
		\includegraphics[width=\linewidth]{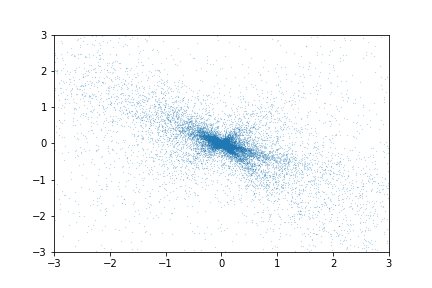}
		\centering{\small (f) Repeat 6}
	\end{minipage}
	\begin{minipage}[c]{0.24\linewidth}
		\includegraphics[width=\linewidth]{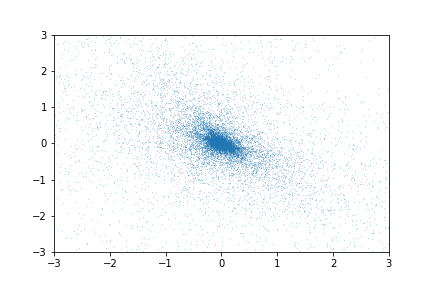}
		\centering{\small (g) Repeat 7}
	\end{minipage}
	\begin{minipage}[c]{0.24\linewidth}
		\includegraphics[width=\linewidth]{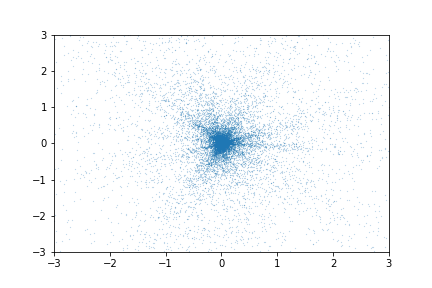}
		\centering{\small (h) Repeat 8}
	\end{minipage}
	\caption{\small Distribution of gradients on MNIST after  epochs 9 projected using different random matrices. 
	}
    \label{fig:mnist_ep9}
\end{figure}

\begin{figure}[htbp]
	\centering
	\begin{minipage}[c]{0.24\linewidth}
		\includegraphics[width=\linewidth]{figures/mnist/epoch59/mnist_gradient_59_0.png}
		\centering{\small (a) Repeat 1}
	\end{minipage}
	\begin{minipage}[c]{0.24\linewidth}
		\includegraphics[width=\linewidth]{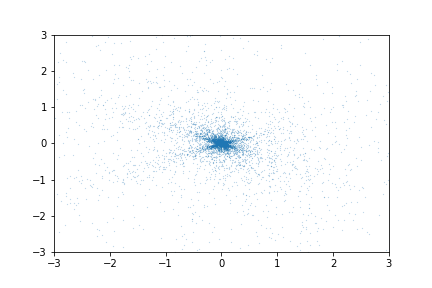}
		\centering{\small (b) Repeat 2}
	\end{minipage}
	\begin{minipage}[c]{0.24\linewidth}
		\includegraphics[width=\linewidth]{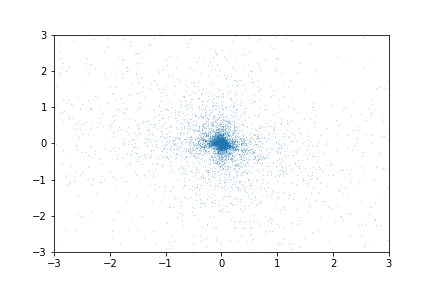}
		\centering{\small (c) Repeat 3}
	\end{minipage}
	\begin{minipage}[c]{0.24\linewidth}
		\includegraphics[width=\linewidth]{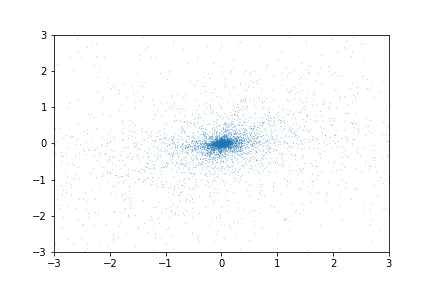}
		\centering{\small (d) Repeat 4}
	\end{minipage}
	\\ 
		\begin{minipage}[c]{0.24\linewidth}
		\includegraphics[width=\linewidth]{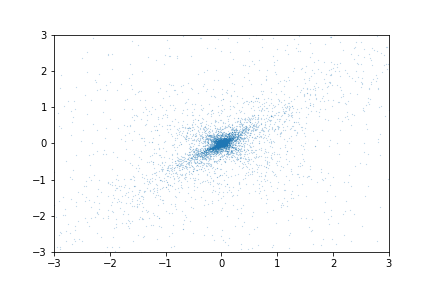}
		\centering{\small (e) Repeat 5}
	\end{minipage}
	\begin{minipage}[c]{0.24\linewidth}
		\includegraphics[width=\linewidth]{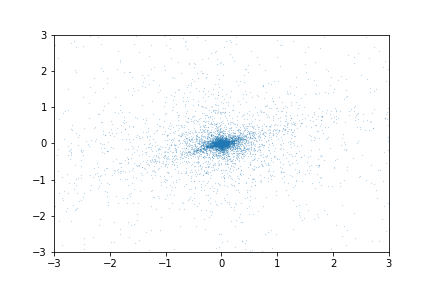}
		\centering{\small (f) Repeat 6}
	\end{minipage}
	\begin{minipage}[c]{0.24\linewidth}
		\includegraphics[width=\linewidth]{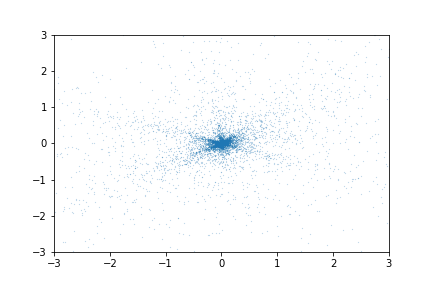}
		\centering{\small (g) Repeat 7}
	\end{minipage}
	\begin{minipage}[c]{0.24\linewidth}
		\includegraphics[width=\linewidth]{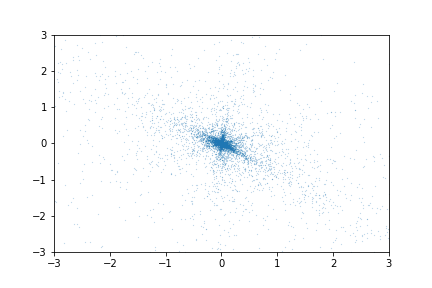}
		\centering{\small (h) Repeat 8}
	\end{minipage}
	\caption{\small Distribution of gradients on MNIST after  epochs 59 projected using different random matrices. 
	}
    \label{fig:mnist_ep59}
\end{figure}

We show the projection of stochastic gradients into 2d space described in Section \ref{sec:experiments} for different projection matrices in Figure \ref{fig:mnist_ep0}-\ref{fig:mnist_ep59}. It can be seen that as the training progresses, the gradient distribution in 2d space tend to be increasingly more symmetric.

\newpage
\section{Evaluation on the probability term}
\begin{figure}[htbp]
	\centering
	\begin{minipage}[c]{0.45\linewidth}
		\includegraphics[width=\linewidth]{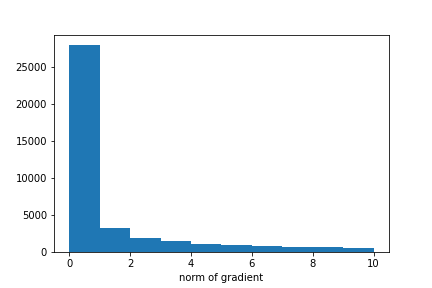}
		\centering{\small (a) Norm of gradients}
	\end{minipage}
	\begin{minipage}[c]{0.45\linewidth}
		\includegraphics[width=\linewidth]{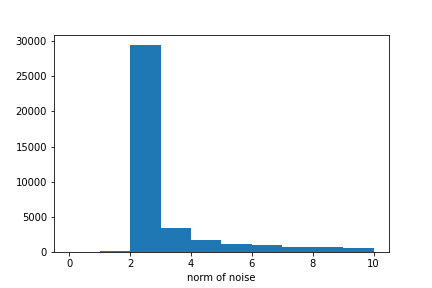}
		\centering{\small (b) Norm of noise}
	\end{minipage}
	\\
		\begin{minipage}[c]{0.45\linewidth}
		\includegraphics[width=\linewidth]{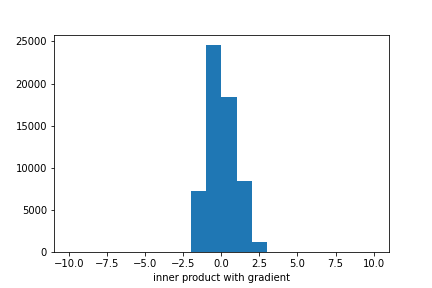}
		\centering{\small (c) Inner product between true gradient and clipped stochastic gradients}
	\end{minipage}
	\begin{minipage}[c]{0.45\linewidth}
		\includegraphics[width=\linewidth]{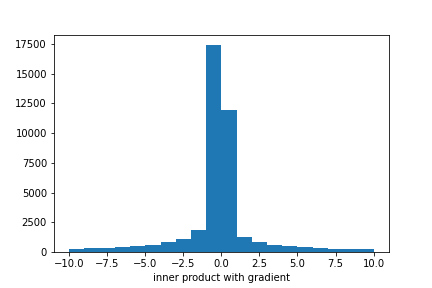}
		\centering{\small (d) Inner product between true gradient and stochastic gradients}
	\end{minipage}
	\caption{\small Distribution of different statistics at epoch 3. 
	}
    \label{fig:mnist_dist_ep3}
\end{figure}

\begin{figure}[htbp]
	\centering
	\begin{minipage}[c]{0.45\linewidth}
		\includegraphics[width=\linewidth]{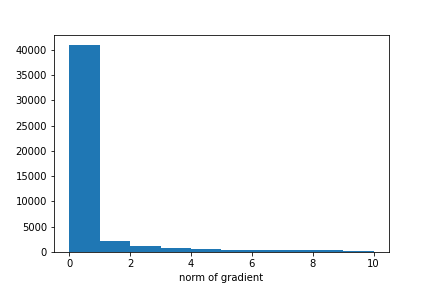}
		\centering{\small (a) Norm of gradients}
	\end{minipage}
	\begin{minipage}[c]{0.45\linewidth}
		\includegraphics[width=\linewidth]{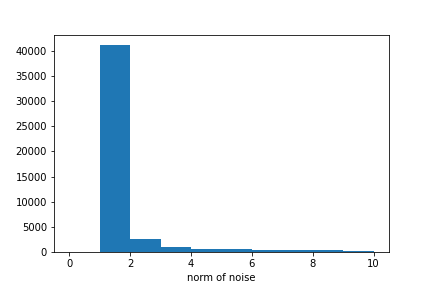}
		\centering{\small (b) Norm of noise}
	\end{minipage}
	\\
		\begin{minipage}[c]{0.45\linewidth}
		\includegraphics[width=\linewidth]{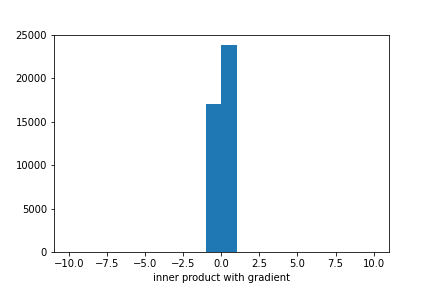}
		\centering{\small (c) Inner product between true gradient and clipped stochastic gradients}
	\end{minipage}
	\begin{minipage}[c]{0.45\linewidth}
		\includegraphics[width=\linewidth]{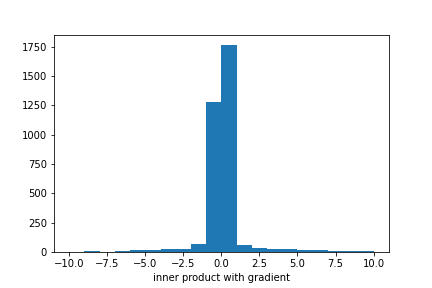}
		\centering{\small (d) Inner product between true gradient and stochastic gradients}
	\end{minipage}
	\caption{\small Distribution of different statistics at epoch 9. 
	}
    \label{fig:mnist_dist_ep9}
\end{figure}

\begin{figure}[htbp]
	\centering
	\begin{minipage}[c]{0.45\linewidth}
		\includegraphics[width=\linewidth]{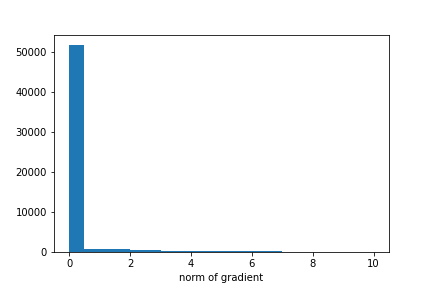}
		\centering{\small (a) Norm of gradients}
	\end{minipage}
	\begin{minipage}[c]{0.45\linewidth}
		\includegraphics[width=\linewidth]{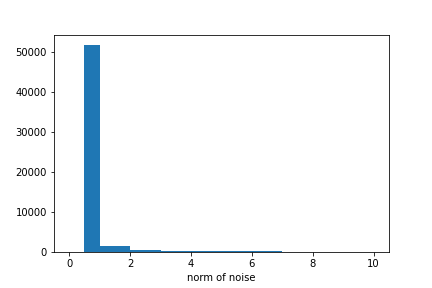}
		\centering{\small (b) Norm of noise}
	\end{minipage}
	\\
		\begin{minipage}[c]{0.45\linewidth}
		\includegraphics[width=\linewidth]{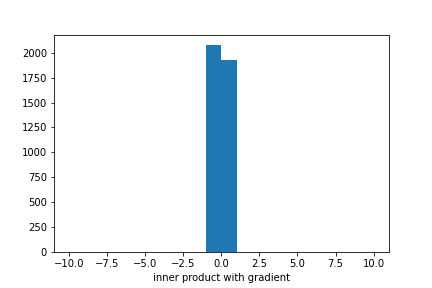}
		\centering{\small (c) Inner product between true gradient and clipped stochastic gradients}
	\end{minipage}
	\begin{minipage}[c]{0.45\linewidth}
		\includegraphics[width=\linewidth]{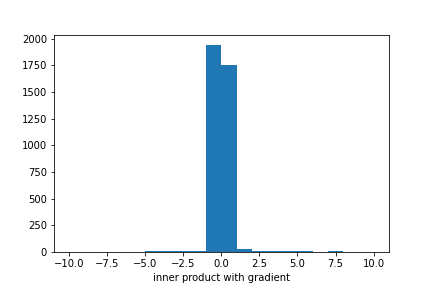}
		\centering{\small (d) Inner product between true gradient and stochastic gradients}
	\end{minipage}
	\caption{\small Distribution of different statistics at epoch 59. 
	}
    \label{fig:mnist_dist_ep59}
\end{figure}

In this section, we evaluate the probability term in Corollary \ref{corl: conv} using a few statistics of the empirical gradient distribution on MNIST. Specifically, at the end of different epochs, we plot histogram of norm of stochastic gradient and norm of noise, along with the inner product between stochastic gradient (and clipped stochastic gradient) and the true gradient. The results are shown in Figure \ref{fig:mnist_dist_ep3}-\ref{fig:mnist_dist_ep59}. One observation is that the norm of stochastic gradients is concentrated around 0 while having a heavy tail. The noise distribution is concentrated around some positive value with a heavy tail, the mode of the noise actually  corresponds to the approximate 0 norm mode of stochastic gradients. As the training progress, the norm of stochastic gradients and the norm of noise are approaching 0. We set clipping threshold to be 1 in the experiment, so actually the probability $\mathbb P (\|\xi_t\| \leq \frac{1}{4}c)$ is 0 for the empirical distribution $p$. When we use a distribution $\tilde p$ with $\mathbb P (\|\xi_t\| \leq \frac{1}{4}c) \geq l$ for some value $l >0$ to approximate $p$, this approximation indeed can create a approximation bias. However, the bias may not be too large since the mode of the norm of noise is not too much bigger than $\frac{c}{4}$. Furthermore, in Corollary \ref{corl: conv} and Theorem \ref{thm: symmetric_descent}, we actually can change $\mathbb P_{\xi_t \sim \tilde p} (\|\xi_t\| \leq \frac{1}{4}c)$ to $\mathbb P_{\xi_t \sim \tilde p} (\|\xi_t\| \leq zc)$ with any $z < 1$ and simultaneously change the $\frac{3}{4}c$ to $(1-z)c$ to make the probability term larger. 

Despite the discussions above, the distribution of norm of stochastic gradients and noise norm combined with the 2d visualization experiments implies the noise on gradient might follow a mixture of distributions with each component being approximate symmetric. Especially one component may correspond to a approximate 0 mean distribution of stochastic gradients. Intuitively this can be true since  each class of data may corresponds to a few variations of stochastic gradients and the gradient noise is observed to be low rank in \cite{LiGZCB20}. We have some discussions in Section \ref{sec: beyond_symmetric} to explain how convergence can be achieved in the cases of symmetric distribution mixtures but it may not be the complete explanation here. Further exploration of gradient distribution in practice is an important question and we leave it for future research.

\newpage
\section{Additional results and discussions on the probability term and the noise adding approach in Section \ref{sec: bias_reduce}} \label{app: noise_adding}
Theorem \ref{thm: scale_noise_main} says that after adding the  Gaussian noise $k\zeta_t$ before clipping, the clipping bias can decrease. In the meantime, the expected decent also decreases because $\mathbb P(\|k\zeta_t\|<\frac{c}{4})$ decreases with $k$. To get a more clear understanding of the theorem, consider $d=1$, then $\mathbb P(\|k\zeta_t\|<\frac{c}{4}) = \textrm{erf}(\frac{c}{4k})$ which decreases with an order of $O(\frac{1}{k})$. This rate is slower than the $O(\frac{1}{k^2})$ diminishing rate of the clipping bias. Thus, as $k$ becomes large, the clipping bias will be negligible compared with the expected descent. This will translate to a {\bf slower} convergence rate with a {\bf better} final gradient bound in convergence analysis. The key idea of adding $k\zeta_t$ before clipping is to "symmetrify" the overall gradient noise distribution. By adding the isotropic symmetric noise $k\zeta_t$, the distribution of the resulting gradient noise $W_t \triangleq \xi_t + k\zeta_t$ will become increasingly more symmetric as $k$ increases. In particular, the total variation distance between the distribution of $W_t$ and $k\zeta_t$ decreases at a rate of $O(\frac{1}{k^2})$ which can be further used to bound the clipping bias. Then, one can apply Theorem \ref{thm: symmetric_descent} to lower bound $ E_{\xi_t=0, \zeta_t}[\langle \nabla f(x_t), g_t \rangle ]$ by letting $\tilde p$ be the distribution of $k\zeta_t$. We believe the lower bounds in Theorem \ref{thm: scale_noise_main} can be further improved when $d > 1$, notice that $\mathbb P(\|k\zeta_t\|<\frac{c}{4})$ tends to decrease fast with $k$ when $d$ being large.

However, we observe $E_{\xi_t \sim {p}, \zeta_t}[\langle \nabla f(x_t), g_t \rangle ]$ decreases with a rate of $O(1/d)$ and $O(1/k)$ in practice  for fixed $\|\nabla f(x_t)\|$ and $\xi_t=0$ (see Table \ref{tab: scale_d_k} 
for $\|\nabla f(x_t)\| = 10$, the expectation $E_{\xi_t =0, \zeta_t}[\langle \nabla f(x_t), g_t \rangle ]$ is evaluated over $10^5$ samples of $\zeta_t \sim \mathcal N(0,I)$). In addition, we found the lower bounds in Theorem \ref{thm: symmetric_descent} are tight up to a constant when $d=1$. {To verify the lower bounds, we considered a 1-dimensional example and choose a symmetric noise $\xi_t \sim \mathcal N (0,1)$ and set $c=1$. Then we compare $\mathbb E_{\xi_t \sim \tilde{p}}[\langle \nabla f(x_t), g_t \rangle ]$ (estimated by averaging $10^5$ samples) with the lower bound in Theorem 2 for different $\|\nabla f(x_t)\|$ and the results are shown in Table \ref{tab: lowerbounds}. Similar result should also hold for $\tilde p( \xi_t)$ being a distribution on a 1 dimensional subspace.} This implies the lower bound can only be improved by using more properties of isotropic distributions like $\mathcal N(0,  I)$ or resorting to a more general form of the lower bounds. We found this to be non-trivial and decide to leave it for future research. 
\begin{table}[h]
  \caption{Scalability of $E_{\xi_t =0, \zeta_t}[\langle \nabla f(x_t), g_t \rangle ]$ w.r.t. $d$ and $k$}
  \label{tab: scale_d_k}
  \centering
\begin{tabular}{ |p{1.5cm}||p{1.5cm}|p{1.5cm}|p{1.5cm}|p{1.5cm}|p{1.5cm}|  }
 \hline
  & $d=1$ & $d=10$ & $d=100$ & $d=1000$ & $d=10000$\\
 \hline
 $k=1$ & 10   & 9.572   & 7.077 &  3.015  &0.995\\
 \hline
  $k=10$ & 6.788   & 2.961   & 0.992 &   0.316 &0.1\\
   \hline
  $k=100$ & 0.758   & 0.316   & 0.098 &   0.032 &0.01\\
     \hline
  $k=1000$ & 0.084   & 0.019    &0.011  &   0.003 &0.001\\
 \hline
\end{tabular}
\end{table}

 The results below verify our lower bound. 
\vspace{-0.25cm}
\begin{table}[h]
  \caption{Verify the lower bounds in Theorem \ref{thm: symmetric_descent}}
  \label{tab: lowerbounds}
  \centering
  \begin{tabular}{ |c|c|c|c|c|c|c| } 
 \hline
  $\|\nabla f(x_t)\|$ & 0.05 & 0.1 & 1 & 2 & 10 & 100  \\ 
   \hline
  $\mathbb E_{\xi_t \sim \mathcal N(0,1)}[\langle \nabla f(x_t), g_t \rangle ]$ & 1.7e-4 & 6.6e-3 &0.612 &1.83 & 10 & 100 \\ 
  \hline
 lower bound & 4e-5 & 2e-3 & 0.148 &0.3 & 1.48 & 14.8 \\ 
   \hline
\end{tabular}
\end{table}
\vspace{-0.2cm}

\end{document}